\makeatletter\def\graphicscache@inhibit{true}\makeatother
\documentclass{article}
\usepackage[utf8]{inputenc}
\usepackage{jfrExamplee}
\usepackage{graphicx}

\usepackage{natbib}
\let\bibhang\relax
\usepackage{apalike}

\usepackage{setspace}

\usepackage{tikz,pgfplots,pgfplotstable}

\usepackage{textcomp}
\usepackage{subcaption}
\usepackage{threeparttable}
\usepackage{xstring}
\usepackage{graphicx}
\usepackage[export]{adjustbox}

\usepackage{tensor}
\usepackage{amsmath}

\usepackage[hidelinks]{hyperref}
\usepackage[capitalise]{cleveref}

\usepackage{siunitx}

\IfFileExists{graphicscache.sty}{\usepackage{graphicscache}}

\usepackage{tikz}
\usetikzlibrary{arrows}
\usetikzlibrary{positioning,calc}
\usetikzlibrary{decorations.pathreplacing}
\usetikzlibrary{decorations.markings}
\usetikzlibrary{fit}
\usetikzlibrary{shapes.callouts}
\usetikzlibrary{shapes.geometric}
\usetikzlibrary{matrix}
\usetikzlibrary{positioning}
\usetikzlibrary{arrows.meta}
\usetikzlibrary{backgrounds}

\usepackage{pgfplots}
\pgfplotsset{compat=newest}
\usepgfplotslibrary{groupplots}
\usepgfplotslibrary{units}
\pgfplotsset{compat=1.15,
  /pgfplots/ybar legend/.style={
    /pgfplots/legend image code/.code={%
       \draw[##1,/tikz/.cd,yshift=-0.25em]
        (0cm,0cm) rectangle (3pt,0.8em);},
   },
}

\newcommand{\T}[2]{\ensuremath{\tensor[^{#1}]{T}{_{#2}}}}
\newcommand{\TT}[2]{\ensuremath{\tensor[^{#1}]{\tilde{T}}{_{#2}}}}

\newcommand{\R}[2]{\ensuremath{\tensor[^{#1}]{R}{_{#2}}}}
\newcommand{\RT}[2]{\ensuremath{\tensor[^{#1}]{\tilde{R}}{_{#2}}}}

\usepackage{booktabs}

\usepackage[tightpage]{preview}

\newenvironment{maybepreview}%
{\noindent\ignorespaces}%
{\par\noindent%
\ignorespacesafterend}
\PreviewEnvironment{maybepreview}

\usepackage[multidot]{grffile}

\usepackage[draft]{fixme}
\fxsetup{theme=color, layout={inline,index}}

\DeclareUnicodeCharacter{FEFF}{}

\usepackage[firstpage=true]{background}
\newcommand\copyrighttext{%
\parbox{\textwidth}{
\footnotesize
Accepted for Field Robotics, Special Issue on MBZIRC 2020: Challenges in Autonomous Field Robotics, to appear 2021.
}
}
\SetBgContents{\copyrighttext}
\SetBgScale{1}
\SetBgColor{black}
\SetBgAngle{0}
\SetBgOpacity{1}
\SetBgPosition{current page.north}
\SetBgVshift{-0.8cm}
\usepackage[bottom]{footmisc}

\title{Team NimbRo's UGV Solution for Autonomous Wall Building and Fire Fighting at MBZIRC 2020}

\author{
Christian Lenz,
Jan Quenzel,
Arul Selvam Periyasamy,
Jan Razlaw,
Andre Rochow,
\\ \textbf{
Malte Splietker,
Michael Schreiber,
Max Schwarz,
Finn Süberkrüb,
and Sven Behnke} \\
Autonomous Intelligent Systems Group \\
University of Bonn \\
Bonn, Germany \\
\texttt{lenz@ais.uni-bonn.de}\\
}

\begin{document}

\maketitle

\begin{abstract}

Autonomous robotic systems for various applications including transport,
mobile manipulation, and disaster response are becoming more and more complex.
Evaluating and analyzing such systems is challenging.
Robotic competitions are designed to benchmark complete robotic systems on complex state-of-the-art tasks.
Participants compete in defined scenarios under equal conditions.
We present our UGV solution developed for the Mohamed Bin Zayed International Robotics
Challenge 2020.
Our hard- and software components to address the challenge tasks of wall building and fire
fighting are integrated into a fully autonomous system.
The robot consists of a wheeled omnidirectional base, a 6\,DoF manipulator arm equipped
with a magnetic gripper, a highly efficient storage system to transport
box-shaped objects, and a water spraying system to fight fires. The robot perceives its environment using 3D LiDAR as well as
RGB and thermal camera-based
perception modules, is capable of picking box-shaped objects and constructing a
pre-defined wall structure, as well as detecting and localizing heat sources in
order to extinguish potential fires. 
A high-level planner solves the challenge tasks using the robot skills.
We analyze and discuss our successful participation during the MBZIRC~2020 finals,
present further experiments, and provide insights to our lessons learned.

\end{abstract}

\section{Introduction}

Autonomous robotic systems have a large potential for future applications, including
transport, mobile manipulation, manufacturing, construction, agriculture,
and disaster-response.
Robots can assist humans in physical demanding or highly repetitive work, such as
industrial construction or transportation tasks. Especially for disaster-response
applications, robots are a useful tool which can be deployed in dangerous environments
to reduce the risk on humans. Autonomous behavior reduces the cognitive load on operators and
allows for deployment in communication-constrained situations.

Evaluating complex robotic systems as a whole under realistic conditions
is challenging. Robotic competitions such as the DARPA Robotics Challenge or the Mohamed
Bin Zayed International Robotics Challenge (MBZIRC) are designed to compare approaches on
a systems level. Participants at such competitions are motivated to advance the state of the art in the domain of interest by working on complex systems designed to solve the specified tasks.
They have the opportunity to test their systems at the competition event
in realistic scenarios and compete against other teams under equal conditions.

\begin{figure}
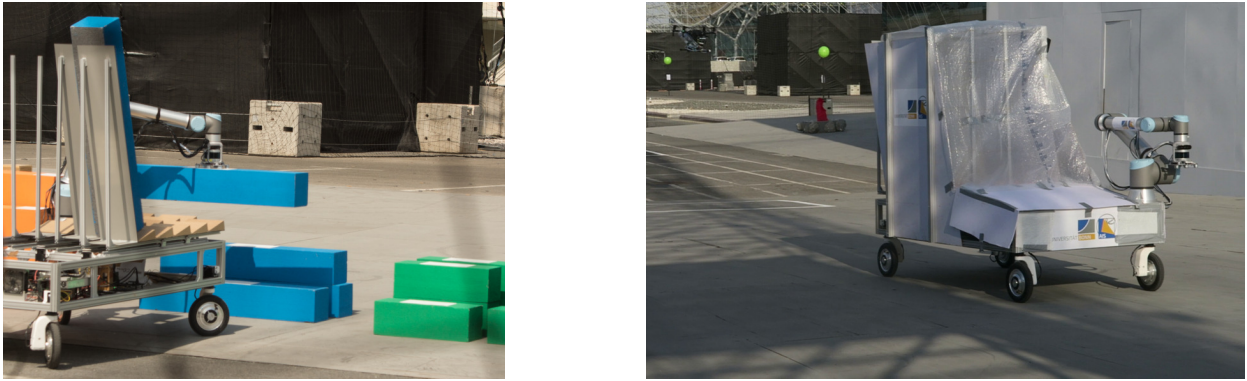

 \centering\begin{maybepreview}%
 \includegraphics[clip,trim=410 250 390 200,height=5.0cm]{images/finn/MBZIRC_2020-65.jpg}\vspace*{-1ex}
 \hfill%
 \includegraphics[clip,trim=200 0 0 0,  height=5.0cm]{images/ch3/bob2.png}\end{maybepreview}%
 \caption{Our UGV Bob during Challenge~2 (left) and Challenge~3 (right) at the MBZIRC 2020 Finals.}
 \label{fig:bob_ch2_ch3}
\end{figure}

In this article, we present our unmanned ground vehicle (UGV) designed to solve wall building and fire fighting tasks in Challenges~2 and 3
of the MBZIRC 2020
finals (see \cref{fig:bob_ch2_ch3}). In addition to describing and analyzing our
integrated system, we discuss lessons learned and detail our technical contribution, including:
A precise laser-based registration and navigation module, a laser-based pose estimation and registration module for
the box-shaped objects, a highly space- and time-efficient box storage system, robust 3D fire localization using thermal cameras, and
an efficient high-level planner for construction.

\section{MBZIRC 2020}

The Mohamed Bin Zayed International Robotics Challenge (MBZIRC) is a biennial international robotics competition. The second edition held in early 2020 consisted
of three different challenges and a Grand Challenge combining all three challenges. In contrast to Challenge~1 allowing only unmanned aerial vehicles (UAV),
Challenges~2 and 3 were designed to be solved by a team of (up to three) UAVs and
an unmanned ground vehicle (UGV). 

\begin{figure}[t]
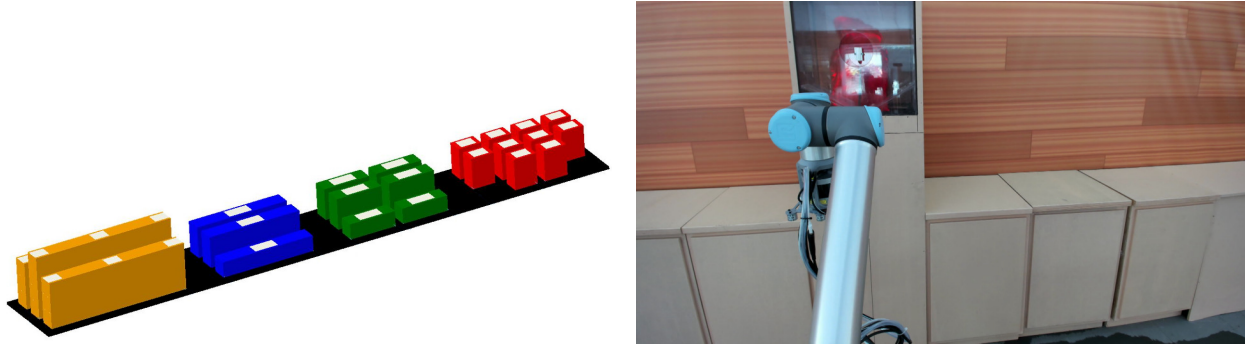

 \centering\begin{maybepreview}%
 \includegraphics[clip, width=.49\linewidth]{images/rules/bob_pile.jpg}
 \hfill%
 \includegraphics[clip, width=.49\linewidth]{images/ch3/bob_teaser.png}\end{maybepreview}%
 \caption{Left: Piles of bricks to be picked up by the UGV in Challenge~2. Right: Simulated indoor fire in Challenge~3.}
 \label{fig:challenges}
\end{figure}

In Challenge~2, the robots had to pick, transport, and place bricks to build a wall.
Four different brick types with 20$\times$20\,cm cross-section were used:
Red (30\,cm length, 1\,kg), green (60\,cm, 1.5\,kg), blue (120\,cm, 1.5\,kg),
and orange (180\,cm, 2\,kg). Each brick had a ferromagnetic patch allowing a magnetic gripper to manipulate them. Each type of robot had a designated pick-up and
place area inside the arena (40$\times$50\,m). \cref{fig:challenges}(left) shows the brick specifications and arrangement of the bricks for the UGV at the beginning of the task.
The robots had to build the first wall segment using only orange bricks. For the
remaining segment, a random blueprint defining the order of the red, green,
and blue bricks was provided some minutes before the competition. Points were granted for correctly placed bricks. The UGV could achieve
 1-4 points per brick (45 bricks in total). The time limit for this
challenge was 25\,min.

Challenge~3 targeted an urban fire-fighting scenario for autonomous robots. A team of up to three UAVs and one UGV had to
detect, approach and extinguish simulated fires around and inside a building. Each fire provided
a 15\,cm circular opening with a heated plate recessed about
10\,cm on the inside. Holes on the outside facade were
surrounded by a ring of propane fire, while indoor fires had
a moving silk flame behind the thermal element. Each fire granted points based on the difficulty reaching the location, scaled by the amount of water (up to 1\,l)
the robots had delivered to extinguish the fire. \cref{fig:challenges}(right) shows the simulated indoor fire to be extinguished by the UGV.

All tasks had to be performed autonomously to achieve
the perfect score. The teams were allowed to call a reset at
any time to bring the robots back to the starting location. Resets did not result in
a point penalty, but no extra time was granted.

\section{Related Work}

\textit{Mobile Manipulation in Industrial Automation:}
Many examples of mobile manipulation robots exist.
For example, \Citet{krueger2016vertical} developed a robotic system for automotive kitting
within the European FP7 project STAMINA. They mount an industrial manipulator arm
on a heavy automated guide vehicle platform.
\Citet{krug2016next} introduce APPLE, a mobile manipulation robot based on
a motorized forklift. The system is capable of autonomous picking and
palletizing using a Kuka iiwa manipulator arm.
In contrast to these systems, our robot has to perform a less-defined task
in an unstructured environment under diverse lighting conditions.

\textit{UGVs for Wall Building:}
The application of robots for wall-building has a long history~\citep{slocum1988blockbot}.
One particularly impressive example is the work of \citet{dorfler2016mobile} who developed a heavy mobile bricklaying robot for the creation of free-form curved walls.
An alternative for creating free-form walls is on-site 3D printing with a large manipulator arm \citep{keating2017toward}.
In contrast to these systems, our robot is able to fetch bricks from piles autonomously.

\textit{UGVs for Disaster Response:}
Our work mostly relates to disaster-response robotics, where protective or otherwise functional structures have
to be built quickly and with minimal human intervention.
The DARPA Robotics Challenge~\citep{krotkov2017darpa} established a baseline
for flexible disaster-response robots. More recent disaster-response robot systems include WAREK-1 \citep{Hashimoto:SSRR2017}, a four legged robot, CENTAURO \citep{Klamt:RAM2019CENTAURO}, a complex robot with an anthropomorphic upper body and a hybrid legged-wheeled base, and E2-DR \citep{Yoshiike:RAM2019}, a humanoid robot.
In comparison to these, our system has a much higher degree of autonomy,
but is more specialized for the task at hand.

\textit{UGVs for Fire Fighting:}
Commercially available ground vehicles for remote fire fighting include the Thermite \citep{Thermite} and LUF \citep{LUF} tracked platforms with steerable nozzle.
They are directly controlled from a safe distance by a fire fighter.   
Cooperative monitoring and detection of forest and wildfires with autonomous teams of UGVs \citep{Ghamry2016icuas} gained
significant attention in recent years \citep{Delmerico2019jfr}.
New challenges arise where the robots have to operate close to structures.
Therefore, UGVs are often equipped with cameras and are remote-controlled by first responders.
In contrast, autonomous execution was the goal for Challenge~3 of MBZIRC 2020.
Team Skyeye \citep{SkyeyeMBZIRC} used a 6-wheeled UGV with color and thermal cameras, GPS and LiDAR.
A map was prebuilt from LiDAR, IMU and GPS data to allow online Monte Carlo localization and path planning with Lazy Theta$^{*}$.
Fires were detected via thresholding on thermal images. The fire location was estimated with an information filter
from either projected LiDAR range measurements or the map.
A water pump for extinguishing was mounted on a pan-tilt unit on the UGV.

Although our general approach is similar to team Skyeye, we rely more heavily upon relative navigation for aiming at the target after initial detection and less on the quality of our map and localization.
In comparison, the placement of the hose on the end-effector of the robot arm on our UGV gives us a better reach.
For detailed descriptions of our UAV/UGV team's participation in Challenges~2 we refer to \citep{ssrr_ch2}.

\textit{MBZIRC 2020:}
\citep{stibinger2021mobile} developed a UGV for the MBZIRC 2020 Challenge~2. The robot uses a 3D LiDAR sensor
for environment sensing and an RGB-D sensor mounted at the endeffector for visual feedback during the grasping process.
Compared to Bob, their solution is much smaller and thus, can transport fewer bricks. Additional color information is
used to detect and localize the bricks.
A UGV for Challenge~3 was developed by \citep{raveendran2020development}. Again, a 3D LiDAR sensor is used for SLAM.
Additional RGB-D camera measurements perceives the environment to detect the target objects (fires) using Darknet.
Our solution in comparison relies on thermal information only for detecting the fires.

\textit{MBZIRC 2017:}
The first edition of the MBZIRC competition also featured a UGV task: Autonomous manipulation of a valve using a wrench selected and grasped by the robot.
While the overall task is different, many subtasks remain similar, such as reliable and fast autonomous navigation, the usage of force sensing
for manipulation, and so on. Our own, winning system for MBZIRC 2017, the UGV Mario \citep{schwarz2019team}, serves as the basis for this work.
It featured a fast omnidirectional base with brushless hub motors, which carried a Universal Robots UR5 arm for mobile manipulation.
Another notable entry are \citet{bilberg2019force}, who attained second place in the UGV challenge with their system built on top of a
four-wheeled skid-steer vehicle. For manipulation, they also employed a UR5 arm, but with very sensitive force sensing capabilities, which
replaced other sensing modalities such as LiDAR---detecting and measuring the manipulation targets by touch alone.

For detailed descriptions of our MAV/UGV team's participation in MBZIRC 2020 Challenge~2, we refer to \citet{ssrr_ch2}.

\section{Hardware Design}

We build our ground robot Bob based on our very successful UGV Mario, which won
the first MBZIRC competition~\citep{schwarz2019team}. We improved the
basis slightly and adapted the manipulator and sensors for the new challenges.
Since 45 bricks had to be picked, transported, and placed
in 25\,min to obtain a perfect score for Challenge~2,
we developed our UGV to store as many bricks as
possible, complying to the size restrictions for the competition.
A much smaller robot platform would have been sufficient to carry the components needed for Challenge~3, but would have increased the
overall competition complexity using an additional system. Thus, we extended Bob
with a water storage for up to 10\,liters and two windscreen washer pumps as our fire extinguish components for Challenge~3.
A 3D LiDAR scanner, two Logitech Brio webcams and a FLIR Lepton 3.5 thermal camera are used as visual sensors for both challenges.

\subsection{Omnidirectional Base}

\begin{figure}
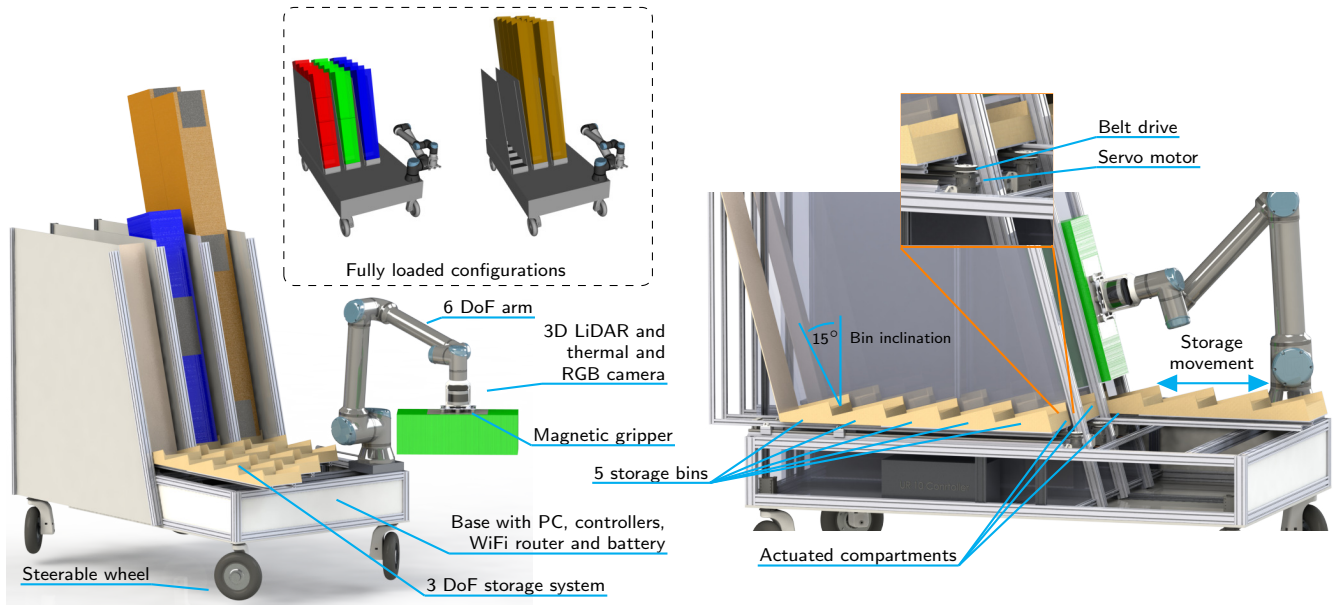

	\centering\begin{maybepreview}%
	﻿\begin{tikzpicture}[
 	font=\sffamily\scriptsize,
    every node/.append style={text depth=.2ex},
	box/.style={rectangle, inner sep=0.5, anchor=west},
	line/.style={cyan, thick}
]

\node[anchor=south west,inner sep=0] (image) at (0,0) {\includegraphics[height=7cm]{images/bob/Bob.png}};
\node[anchor=south west,inner sep=0] (image) at (6.3,4.9) {\includegraphics[height=3cm]{images/bob/storage_orange.png}};
\node[anchor=south west,inner sep=0] (image) at (3.8,4.9) {\includegraphics[height=3cm]{images/bob/storage_color.png}};

\node[box, align=center](rviz) at(4.5,4.5){Fully loaded configurations};
\draw[rounded corners, dashed] (3.7,4.3) rectangle (8.6,8.0);

\node[box, align=right, anchor=east](laser_scanner) at(8.8,3.4){3D LiDAR and\\thermal and\\ RGB camera};
\draw[line](laser_scanner.south west)--(laser_scanner.south east);
\draw[line](laser_scanner.south west)--(6.4,2.9);

\node[box](arm) at(5.8,4.0){6 DoF arm};
\draw[line](arm.south west)--(arm.south east);
\draw[line](arm.south west)--(5.6,3.8);

\node[box](wheel) at(0.2,0.45){Steerable wheel};
\draw[line](wheel.south west)--(wheel.south east);
\draw[line](wheel.south east)--(2.7,0.25);

\node[box, align=right,, anchor=east](base) at(8.8,1.0){Base with PC, controllers,\\WiFi router and battery};
\draw[line](base.south west)--(base.south east);
\draw[line](base.south west)--(4.4,1.45);

\node[box, align=right, anchor=east](gripper) at (8.9,2.3){Magnetic gripper};
\draw[line](gripper.south west)--(gripper.south east);
\draw[line](gripper.south west)--(6.2,2.6);

\node[box, align = left,  anchor=east](storage) at (8.0,0.3){3 DoF storage system};
\draw[line](storage.south west)--(storage.south east);
\draw[line](storage.south west)--(3.1,1.9);

\node[anchor=south west,inner sep=0] (image) at (9.3,1.0) {\includegraphics[clip, trim=0 200 0 250, height=4.5cm]{images/bob/Storage.png}};

\draw[>=triangle 45, <->, cyan, thick] (15.3,3.0) -- (16.8,3.0);
\node[box, align = center] () at (15.45, 3.4) {Storage\\ movement};
\node[box](bin) at(7.8,1.8){5 storage bins};
\draw[line](bin.south west)--(bin.south east);
\draw[line](bin.south east)--(10.6,2.51);
\draw[line](bin.south east)--(11.3,2.5);
\draw[line](bin.south east)--(12.05,2.48);
\draw[line](bin.south east)--(12.8,2.47);
\draw[line](bin.south east)--(13.5,2.46);

\node[box](comp) at(10.0,0.7){Actuated compartments};
\draw[line](comp.south west)--(comp.south east);
\draw[line](comp.south east)--(14.1, 2.5);
\draw[line](comp.south east)--(14.45, 2.7);
\draw[line](comp.south east)--(14.8, 2.6);

\node[anchor=south west,inner sep=0, draw, orange, thick] (close) at (11.9,4.8) {\includegraphics[clip, trim=0 0 0 0, height=2.0cm]{images/bob/StorageCloseUp01.png}};
\draw[line, orange](close.south west)--(14.0, 2.6);
\draw[line, orange](close.south east)--(14.2, 2.6);

\node[box](motor) at(14.5,6.0){Servo motor};
\draw[line](motor.south west)--(motor.south east);
\draw[line](motor.south west)--(13.0, 5.7);

\node[box](belt) at(14.5,6.4){Belt drive};
\draw[line](belt.south west)--(belt.south east);
\draw[line](belt.south west)--(12.9, 5.85);

\draw[line](11.1,2.7)--(11.1, 3.9);
\draw[line](11.1,2.7)--(10.55, 3.8);

\draw[line](11.015,3.8) arc (90:125:0.6);
\node[box, black] () at(10.7,3.6) {{\tiny 15$^{\tiny \circ}$}};
\node[box, black] () at(11.2,3.6) {{\tiny Bin inclination}};

\end{tikzpicture}
\end{maybepreview}%
	\caption{Left: UGV hardware design. Top: The storage system is capable of holding either 10 orange bricks (left)
	or all remaining bricks (20 red, 10 green, 5 blue) (right). Right: Actuated storage system.}
	\label{fig:bob}
\end{figure}

The base has a footprint of 1.9$\times$1.4\,m to provide enough space for our storage
system. It rolls on four direct-drive brushless DC hub motors, controlled by two ODrive driver boards.
Since the motors were originally intended for hover boards, i.e. personal conveyance devices,
they have enough torque to accelerate the approx. 90\,kg UGV. To achieve
omnidirectional movement, we coupled each wheel with a Dynamixel H54-200-S500-R servo which
rotates it around the vertical axis ($0^{\circ}$ caster angle).
The developed base supports driving speeds of up to 4\,m/s as well as very slow speeds $<$0.05\,m/s, which allows
precise positioning relative to perceived objects for manipulation tasks.

MBZIRC 2020 Challenge~2 required to manipulate objects of up to 1.80\,m length
and to stack them to a total height of 1\,m. Instead of the UR5 mounted on our
2017 robot, we used a UR10e 6-DoF robotic
arm, which gives us the benefit of a larger workspace (1.30\,m) and
enough payload capability (10\,kg) to carry the gripper, the sensors, and the
bricks (up to 2\,kg). In addition, the arm is equipped with a force-torque sensor at the last link, which was used for
detecting contact between the gripper or the attached brick and the environment. We adapted the arm controller
to work with UGV battery power.

The UGV is equipped with a standard ATX mainboard with a quad-core
Intel Core i7-6700 CPU and 64\,GB RAM. A discrete GPU is not installed but could be
easily added if necessary for future applications. The whole system is powered by an
eight-cell LiPo battery with 20\,Ah and 29.6\,V nominal voltage. This allows the
robot to operate for roughly one hour, depending on the task.
A microcontroller (Arduino Nano 3) is used to control eight electromagnets and two
windshield washer pumps for Challenge~3.

A Logitech Brio camera is mounted at the rear-top of the robot and is used for
operator situation awareness only. Originally a GPS \& IMU module was added to improve the localization. It turned out that the
wheel odometry and local localization using the 3D LiDAR sensor were good enough for the precise global positioning
required in Challenge~3, as demonstrated by the successful fire extinguish run. Thus, the GPS module was not used and finally removed.
In Challenge~2, only wheel odometry was used for localization purposes. Precise positioning relative to the relevant objects
was achieved using the perception modules which tolerated misalignments of at least 1\,m.

For Challenge~3, we equipped the base with two water tanks containing 5\,liters each, two windscreen water pumps controlled by an Arduino Nano v3
microcontroller and protected the base against splashing water.

\subsection{Multi Purpose End-effector}

\begin{figure}
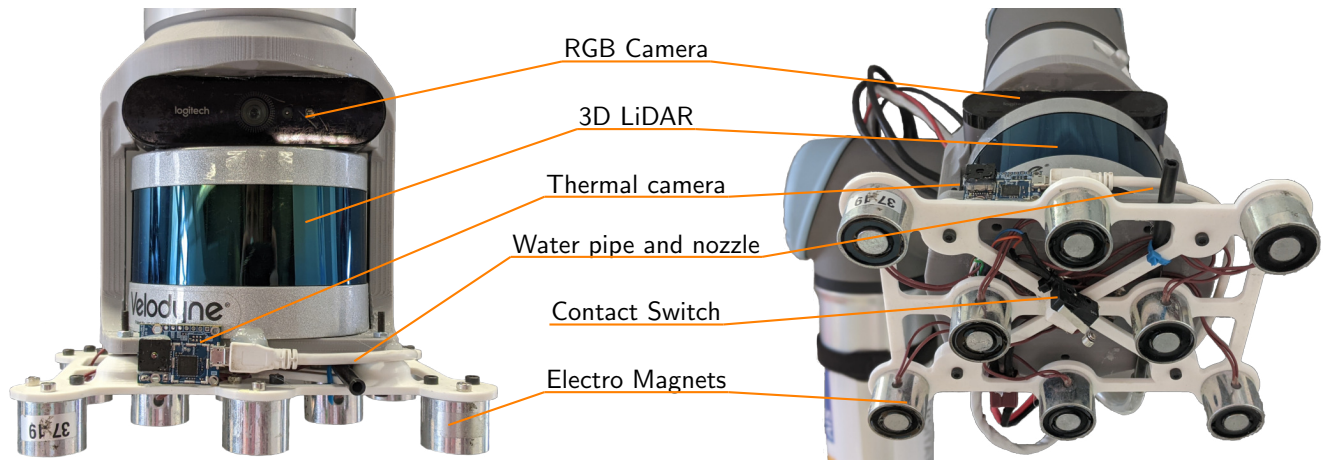

	\centering\begin{maybepreview}%
	﻿\begin{tikzpicture}[
 	font=\sffamily\normalsize,
    every node/.append style={text depth=.2ex},
	box/.style={rectangle, inner sep=0.5, anchor=west},
	line/.style={orange, thick}
]

\node[anchor=south west,inner sep=0] (image) at (0,0) {\includegraphics[height=6cm]{images/bob/gripper/front_transparent.png}};
\node[anchor=south west,inner sep=0] (image) at (10,0) {\includegraphics[height=6cm]{images/bob/gripper/bottom_transparent.png}};

\node[box, align=right, anchor=center](rgb) at(8.4,5.5){RGB Camera};
\draw[line](rgb.south west)--(rgb.south east);
\draw[line](rgb.south west)--(4.0,4.6);
\draw[line](rgb.south east)--(13.5,4.85);

\node[box, align=right, anchor=center](laser_scanner) at(8.4,4.6){3D LiDAR};
\draw[line](laser_scanner.south west)--(laser_scanner.south east);
\draw[line](laser_scanner.south west)--(4.0,3.2);
\draw[line](laser_scanner.south east)--(14.0,4.2);

\node[box, align=right, anchor=center](thermal) at(8.4,3.7){Thermal camera};
\draw[line](thermal.south west)--(thermal.south east);
\draw[line](thermal.south west)--(2.6,1.6);
\draw[line](thermal.south east)--(12.7,3.7);

\node[box, align=right, anchor=center](base) at(8.4,2.9){Water pipe and nozzle};
\draw[line](base.south west)--(base.south east);
\draw[line](base.south west)--(4.65,1.35);
\draw[line](base.south east)--(15.27,3.65);

\node[box, align = left,  anchor=center](storage) at (8.4,2.0){Contact Switch};
\draw[line](storage.south west)--(storage.south east);
\draw[line](storage.south east)--(14.0,2.2);

\node[box, align=right, anchor=center](gripper) at (8.4,1.1){Electro Magnets};
\draw[line](gripper.south west)--(gripper.south east);
\draw[line](gripper.south west)--(5.9,0.5);
\draw[line](gripper.south east)--(12.0,0.8);

\end{tikzpicture}
\end{maybepreview}%
	\vspace*{-2ex}
	\caption{Front (left) and bottom view (right) of our end-effector which incorporates multiple vision sensors, magnets for grasping purposes, a contact switch and a nozzle for water dispensation.}
	\label{fig:hardware_gripper}
\end{figure}

Bob's gripper was designed to feature all required components while keeping the size minimal to reduce the collision potential.
\cref{fig:hardware_gripper} shows the gripper mounted to the UR10e arm.

For perceiving the environment, a Velodyne VLP-16 3D LiDAR sensor and a Logitech Brio RGB camera are mounted inside
the gripper. This allows to move both sensors to the required view poses during perception. The LiDAR sensor is used to perceive
the brick poses in Challenge~2 (see \cref{sec:brick_detection}) and for navigation in Challenge~3 (see \cref{sec:laser_navigation}).
The Logitech Brio RGB camera is used in Challenge~2 only to detect the wall marker (see \cref{sec:wall_marker}).

Challenge~2 required a mechanism to pick up bricks which featured a ferromagnetic part.
Since the ferromagnetic parts of the
bricks are very thin (approx. 0.6\,mm), we decided to use eight smaller electromagnets with 60\,N holding force (measured using a
1\,cm thick metal) to distribute the contact surface as much as possible while keeping the total gripper size minimal.
The magnets are strong enough to manipulate the up to 2\,kg heavy bricks.
The gripper includes a contact switch to detect if a brick is securely grasped.

For Challenge~3, a FLIR Lepton 3.5 thermal camera was added to detect the heat source (see \cref{sec:thermal_detection}) as well as
a hose and carbon fiber nozzle at the gripper for controlled water spraying.
All components are mounted such that bricks can be grasped collision free, the gripper is not prone to self-collisions with
the manipulator, and the sensors are not obstructed.

\subsection{Storage System}

The storage system was designed to provide space for as many bricks as possible
storing the bricks securely in a known pose on the robot which is reachable by the arm (see \cref{fig:bob}).
To achieve these requirements, the storage system has three individual actuated storage compartments.
Each compartment has five
bins to store bricks. The ground plate of each bin is 20.5\,cm wide, 20\,cm
long and is mounted inclined 15$^\circ$ backwards. This inclination forces the
bricks to slide in a known pose inside the storage system even if the bricks are grasped
imprecise. Therefore, we do not need an additional perception system to perceive
the current pose of the stored bricks.

Side walls hold the bricks in place during UGV movements.
The walls are 110\,cm high which is sufficient to
hold the largest bricks (180\,cm long) in place. Furthermore, this system allows to
stack multiple small bricks (up to 4 red bricks, and up to 2 green bricks) to
increase the number of bricks to be stored in the system. Overall, the system is
capable to store either all large bricks (10 orange), or all remaining bricks
(20 red, 10 green, 5 blue, see \cref{fig:bob}).
The bricks are stored sorted by type. This allows to unload any brick type without needing to re-sort the storage system.
Since the storage system exceeds the workspace of the
UR10e, each compartment can be moved horizontally (using a Dynamixel Pro L42-10-S300-R and a linear belt drive)
to put the desired bin in reach of the arm.

\section{High-level Control}

Instead of starting software development from scratch, we extended and further improved several components from our MBZIRC 2017 entry.
We build our software on top of the ROS middleware~\citep{ROS}, the de facto standard framework for robotic applications.

Finite state machines (FSM) are a standard tool for definition and control of robot behavior. For relatively constrained tasks such
as the ones defined by MBZIRC, they allow fast construction of behaviors. Instead of working with standard ROS tools such as
\texttt{SMACH}, a Python-based FSM framework, we decided to develop our own \texttt{nimbro\_fsm2}\footnote{\url{https://github.com/AIS-Bonn/nimbro_fsm2}} library with a focus on
compile-time verification. Since testing time on the real robots is limited and simulation can only provide a rough
approximation of the real systems, it is highly likely that the robot will encounter untested situations during a competition run.
We trade the ease-of-use of dynamically typed languages and standard frameworks against compile-time guarantees to guard against
unexpected failures during runtime.

The \texttt{nimbro\_fsm2} library supports FSM definition in C++. The entire state graph is known
at compile time through the use of C++ metaprogramming features. The library also automatically publishes monitoring data
so that a human supervisor can see the current status. An accompanying GUI displays this data and can trigger manual
state transitions, which is highly useful during testing.

We developed two separate FSMs as high-level controllers for both challenges. Common behavior such as navigation, or arm movement
states was used for both challenges.

\subsection{Challenge~2}
\label{sec:ch2fsm}
The high-level controller for Challenge~2 consists of an FSM
generating the robot actions, a database to keep track of
every brick relevant for the UGV, and an algorithm computing the time-optimal
strategy for a given build order.

\begin{figure}[t]
 \centering\begin{maybepreview}%
 \includegraphics[clip, trim=2 0 0 0, width=\linewidth]{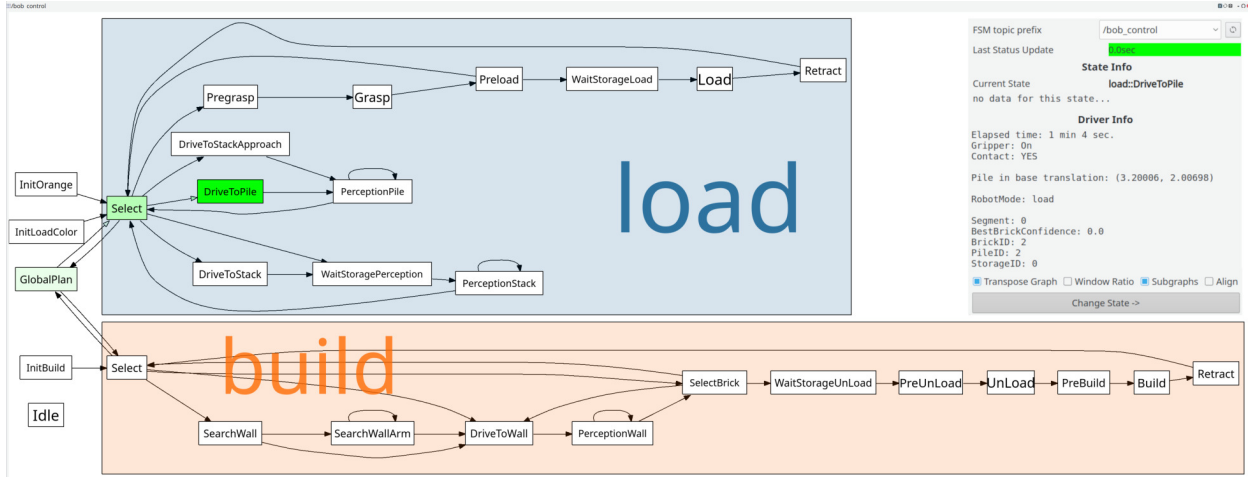}\end{maybepreview}%
 \caption{Graphical user interface showing all states of the FSM developed for Challenge~2. User defined information are shown in the status bar on the right.
 The active state is highlighted in green. The operator can perform state transitions manually using this GUI.}
 \label{fig:fsm_ch2}
\end{figure}

The FSM includes 30 different states for locomotion, manipulation,
perception, storage logistics, and fallback mechanisms (see \cref{fig:fsm_ch2}).
After executing an action, the resulting database and FSM state
was stored to enable quick recovery after a reset.

Since the UR10 arm is very precise, we can manipulate multiple bricks from a stationary
position with perceiving the environment just once. Our overall strategy was
to minimize locomotion between different positions in front
of the piles and the wall. Thus, the plan was to pick up all orange bricks and bring them
to the wall at once. After successfully building the orange wall segment, the UGV
collects all remaining bricks and builds the second wall segment.
Whereas the orange wall segment (two stacks of five bricks each) can be built from
two predefined positions relative to the wall marker, the build order of the second wall segment highly
depends on the supplied blueprint and can be optimized to minimize the number
of locomotion actions.

We implemented a backtracking algorithm to find the optimum build order. To make
this approach feasible regarding run-time, we only consider building the wall from
left to right, but allow starting the next layer before finishing the previous.
Let the longer wall axis (from left to right) be denoted as the x-axis. First,
we compute the set of possible place positions by $P = \{x_i + t_x | x_i = \text{center of brick $i$}\}$.
The place pose is shifted by the arm reach $t_x = 0.675\text{\,m}$ to place the
robot such that the number of brick placement poses in reach is maximized.
Due to the wall structure, we have $ 7 \leq |P| \leq 35 $. We now enumerate over all possible
ordered sequences $S \subseteq P$. For each $p_i \in S$, we build all bricks which meet the following
criteria: 1.) The brick was not built already, 2.) the brick is in reach based on the position $p_i$,
 3.) the brick is fully supported by the ground or previously built bricks, and 4.) the left adjacent brick was built.

\noindent $S = (p_1, p_2, \dots)$ is a valid solution if all bricks are built. We search
for the optimal solution with $|S|$ and $d_S$ minimal,
where $d_S = \sum_{i=2}^{|S|} |p_i - p_{i-1}|$, i.e. the shortest path to
traverse between all building positions.
Pruning sub-trees is used to accelerate the algorithm. Since the optimal strategy
only depends on the desired wall structure, it has to be executed just once before
the competition run starts and thus longer runtimes compared to a greedy strategy
can be justified (see \cref{sec:evaluation}).

\subsection{Challenge~3}

\begin{figure}[h]
  \centering\begin{maybepreview}%
  \resizebox{0.7\linewidth}{!}{\begin{tikzpicture}
[content_node/.append style={minimum size=1.5em,minimum width=6em,draw,align=center,rounded corners,scale=0.65},
label_node/.append style={scale=0.5},
group_node/.append style={dotted,align=center,rounded corners,inner sep=1em,thick},>={Stealth[inset=0pt,length=4pt,angle'=45]}]

\definecolor{red}{rgb}     {0.5,0.0,0.0}
\definecolor{green}{rgb}   {0.0,0.5,0.0}
\definecolor{blue}{rgb}    {0.0,0.0,0.5}
\definecolor{grey}{rgb}    {0.5,0.5,0.5}

\draw[thick, rounded corners, grey!20!white,fill] (-4.0,0.6) -- (4.75,0.6) -- (4.75,4.7) -- (-4.0,4.7) -- cycle;
\draw[thick, rounded corners, grey!20!white,fill] (-4.0,-1.75) -- (4.75,-1.75) -- (4.75,0.0) -- (-4.0,0.0) -- cycle;

\node(Operator)[content_node,fill=green!15!white] at (0.0,3.0) {Operator};
\node(Operator_Camera)[content_node,fill=green!15!white] at (0.0,4.0) {Operator\\Camera};
\node(Thermal_Camera)[content_node,fill=green!15!white] at (3.0,4.0) {Thermal\\Camera};
\node(LIDAR)[content_node,fill=green!15!white] at (-3.0,4.0) {LiDAR};
\node(Thermal_Detection)[content_node,fill=blue!15!white] at (3.0,3.0) {Thermal\\Detection};
\node(Laser_Localization)[content_node,fill=blue!15!white] at (-3.0,3.0) {Laser\\Localization};

\node(State_Machine)[content_node,fill=blue!15!white] at (0.0,1.8) {State\\Machine};
\node(Trajectory_Generation)[content_node,fill=blue!15!white] at (-3.0,1) {Trajectory\\Generation};

\node(Motor_Driver)[content_node,fill=red!15!white] at (-3.0,-0.875) {UGV Motor\\Driver};
\node(Arm_Controller)[content_node,fill=red!15!white] at (0.0,-0.5) {Arm\\Controller};
\node(Arm)[content_node,fill=red!15!white] at (0.0,-1.25) {Arm};
\node(UC)[content_node,fill=red!15!white] at (3.0, -0.5) {Arduino};
\node(Pump)[content_node,fill=red!15!white] at (3.0,-1.25) {Water\\Pump};

\draw[->, thick] (Thermal_Camera) -- node[label_node,midway,left] {\SI{8.6}{\hertz}} node[label_node,midway,right] {Image} (Thermal_Detection);
\draw[->, thick] (Thermal_Detection) -- node[label_node,midway,left] {\SI{10}{\hertz}} node[label_node,midway,right,align=left] {3D~Position\\3D~Orientation} ++(0, -1) |- (State_Machine.10);

\draw[->,thick] (LIDAR) -- (Laser_Localization);
\draw[->, thick] (Laser_Localization) -- node[label_node,midway,left] {\SI{10}{\hertz}} node[label_node,midway,right,align=left] {3D~Position\\3D~Orientation} ++(0, -1) |- (State_Machine.170);

\draw[->, thick] (State_Machine.190) -- node[label_node,midway,below] {\SI{50}{\hertz}\qquad3D~Target~Pose} (State_Machine.190 -| Trajectory_Generation.90) -- (Trajectory_Generation.90);
\draw[->, thick] (Trajectory_Generation) -- node[label_node,midway,left,yshift=0.45cm] {\SI{50}{\hertz}} node[label_node,midway,right,text width=2cm,yshift=0.45cm] {Velocity} (Motor_Driver);

\draw[->, thick] (Operator_Camera) -- node[label_node,midway,left] {\SI{30}{\hertz}} node[label_node,midway,right] {Image} (Operator);
\draw[->, thick] (Operator) -- node[label_node,midway,right, text width=2.0cm, yshift=0.1cm] {Start/Stop Command} (State_Machine);

\draw[->, thick] (State_Machine.south) -- node[label_node,midway,right,text width=2.0cm, yshift=-0.70cm] {Endeffector Pose} (Arm_Controller); \draw[->, thick] (Arm_Controller) -- (Arm);

\draw[->, thick] (State_Machine.350) -| node[label_node,midway,right,text width=1.7cm, yshift=-2.75cm] {Trigger Command} (UC);
\draw[->, thick] (UC) -- (Pump);

\node(ROS_Group_Label)[label_node,anchor=north west] at (-4.0,4.7) {\textbf{Onboard Computer}};
\node(UGV_Group_Label)[label_node,anchor=south west] at (-4.0,-1.75) {\textbf{UGV Hardware}};
   
\end{tikzpicture}
}\end{maybepreview}%
  \caption{UGV system overview for Challenge~3. Green: Onboard Sensors and the operator input. Blue: Software components. Red: Onboard hardware components.}
  \label{fig:ugv_system}
\end{figure}
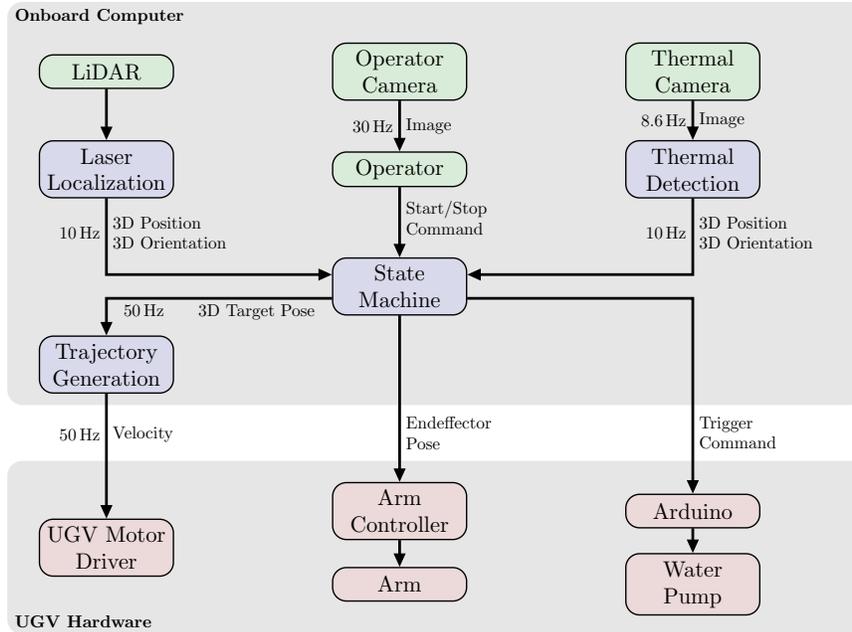

The finite state machine for solving Challenge~3 consists of two main phases. First, the robot has to locate and approach the fire. In the second
phase, the robot aims the nozzle precisely at the fire and starts ejecting water to complete the fire extinguishing task.

\begin{figure}[h]
  \centering\begin{maybepreview}%
  \resizebox{0.8\linewidth}{!}{\begin{tikzpicture}[font=\sffamily,on grid,>={Stealth[inset=0pt,length=4pt,angle'=45]}]
\tikzset{every node/.append style={node distance=3.0cm}}
\tikzset{terminal_node/.append style={minimum size=1.5em,minimum height=3em,minimum width={width("Search Point")+0.2em},draw,align=center,rounded corners,scale=0.65}}
\tikzset{content_node/.append style={minimum size=1.5em,minimum height=3em,minimum width={width("Search Point")+0.2em},draw,align=center,scale=0.65,fill=blue!15!white}}
\tikzset{label_node/.append style={scale=0.5, near start}}
\tikzset{group_node/.append style={align=center,rounded corners,inner sep=1em,thick}}
\tikzset{decision_node/.append style={align=center,scale=0.5,shape aspect=1.5,minimum width=7.9em,minimum height=5.4em,diamond,draw,fill=yellow!25!white,font=\sffamily\normalsize,node distance=3.9cm}}

\definecolor{red}{rgb}     {0.5,0.0,0.0}
\definecolor{green}{rgb}   {0.0,0.5,0.0}
\definecolor{blue}{rgb}    {0.0,0.0,0.5}
\definecolor{grey}{rgb}    {0.5,0.5,0.5}

\draw[thick, rounded corners, grey!20!white,fill] (1, 3.2) -- (4.75, 3.2) -- (4.75, -0.8) -- (1, -0.8) -- cycle;
\draw[thick, rounded corners, grey!20!white,fill] (4.95, 3.2) -- (11.1, 3.2) -- (11.1, -0.8) -- (4.95, -0.8) -- cycle;

\node(start)[terminal_node,fill=red!15!white] at (0, 0) {Start};
\node(enter_building)[content_node, above of=start] {Enter Building};
\node(has_next_fire)[decision_node, right of=enter_building] {Has Next\\Location?};
\node(drive_to_fire)[content_node, below of=has_next_fire] {UGV Drive to\\Location};
\node(perform_search_motion)[content_node, right of=drive_to_fire] {Perform Arm\\Search Motion};
\node(has_thermal_detection)[decision_node, above of=perform_search_motion] {Has Thermal\\Detection?};

\node(aim_hose)[content_node, right of=has_thermal_detection] {Aim Hose};
\node(start_pump)[content_node, below of=aim_hose] {Start Pump};
\node(phase1)[decision_node, right of=start_pump] {Phase 1\\Complete?};
\node(spray_pattern)[content_node, right of=phase1] {Spray in\\Pattern};
\node(phase2)[decision_node, above of=spray_pattern] {Phase 2\\Complete?};
\node(stop_pump)[content_node, left of=phase2] {Stop Pump};
\node(stop)[terminal_node,fill=red!15!white,above of=has_next_fire] {Stop};

\draw[->, thick] (start) -- (enter_building);
\draw[->, thick] (enter_building) -- (has_next_fire);

\draw[->, thick] (has_next_fire) -- node[label_node,right] {Yes} (drive_to_fire);
\draw[->, thick] (has_next_fire) -- node[label_node,right] {No} (stop);
\draw[->, thick] (has_thermal_detection) -- node[label_node,above] {Yes} (aim_hose);
\draw[->, thick] (drive_to_fire) -- (perform_search_motion);
\draw[->, thick] (has_thermal_detection) -- node[label_node,above] {No} (has_next_fire);
\draw[->, thick] (has_thermal_detection) -- (aim_hose);
\draw[->, thick] (perform_search_motion) -- (has_thermal_detection);

\draw[->, thick] (aim_hose) -- (start_pump);
\draw[->, thick] (start_pump) -- (phase1);
\draw[->, thick] (phase1) -- node[label_node,above] {Yes} (spray_pattern);
\draw[->, thick] (phase1.north) -- node[label_node,right, midway] {No} ++(0, 0.4) -- ++(-1.1, 0) |- (phase1);
\draw[->, thick] (spray_pattern) -- (phase2);
\draw[->, thick] (phase2) -- node[label_node,above] {Yes} (stop_pump);
\draw[->, thick] (phase2.east) -- node[label_node,above, midway] {No} ++(0.4, 0) -- ++(0, -1.2) -| (phase2);
\draw[->, thick] (stop_pump) -- ++(0, 0.8) -- +(-5, 0) -- (has_next_fire);

\node[scale=0.65, anchor=north east] at (4.5, 3.2) {\textbf{Search Fire}};
\node[scale=0.65, anchor=north east] at (11, 3.2) {\textbf{Extinguish}};
\end{tikzpicture}
}\end{maybepreview}%
  \caption{Flowchart of the Challenge~3 UGV state machine.}
  \label{fig:ugv_state_machine}
\end{figure}
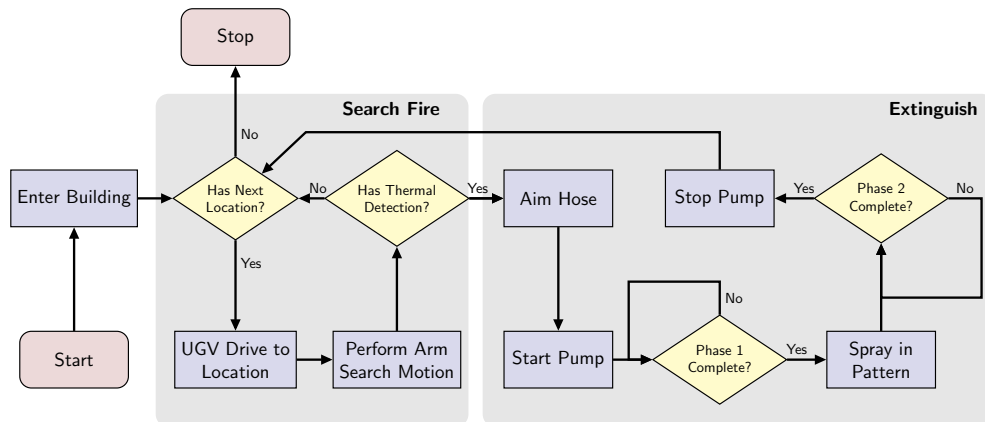

\cref{fig:ugv_system} gives an overview of our system. Details for the central
state machine are shown in \cref{fig:ugv_state_machine}. The UGV navigates from
the start zone to the inside of the building using a set of static waypoints.
In addition to the known arena size, robot start location, and shape of the building,
a map of the building and its location was created during the rehearsal runs and
thus known to the robot. The scenario specifies exactly two static fire locations
within the kitchen, one of which is burning, which we utilized by driving the UGV
to pre-determined poses roughly in front of the fire and performing a rectangular
search motion with the arm. Bob could have been equipped with a more
complex search behavior to detect the fire location without
any prior knowledge of the building's inside. We decided for the much
simpler approach using the generated offline map to increase robustness and execution speed of our system.
When a heat source is detected, the robot starts the
extinguishing procedure. If not, it proceeds to the next location.

The hose is aimed so that the water jet will directly hit the heat element. The water flow trajectory was determined experimentally for a fixed distance between the nozzle and the heat element, resulting in a predefined arm spray pose relative to the heat element. Since the relative position estimate of the heat source described in \cref{sec:thermal_detection} might differ from our calibration and may also vary with distance, the aiming process is performed in multiple \SI{10}{\centi\meter} arm motions to incorporate new measurements. To further increase chances of successfully extinguishing, the actual water spraying is split into two phases. For the first half of the assigned water the hose is kept steady. In the second phase, the remaining water is sprayed in an hourglass-shaped pattern by tilting the endeffector. During the Grand Challenge run the first phase already delivered a sufficient amount of water to score maximum points.
Even though we never observed the thermal detector producing any false-positives during our runs, the robot will continue proceeding to the second fire after extinguishing the first one. The water capacity is more than sufficient with approximately 4\,liters per fire.

\section{Perception}

Autonomous robots base their actions on sensor measurements for perceiving their environment.
We developed multiple perception modules to observe the relevant objects in the robot's vicinity.
For Challenge~2, mainly the 3D laser measurements were used to detect and localize the box-shaped bricks.
Only the marker indicating the desired wall construction pose was invisible for the 3D laser scanner due to
it's flat (approx. 3\,mm) shape. Instead, we used RGB camera images to localize the marker.
Challenge~3 required precise localization next to and inside a known mockup house.
Again, the 3D laser scanner was a perfect sensor for perceiving the building.
The heat source of the simulated fire was observed using a thermal camera.

\subsection{Brick and Wall Perception}

\label{sec:brick_detection}

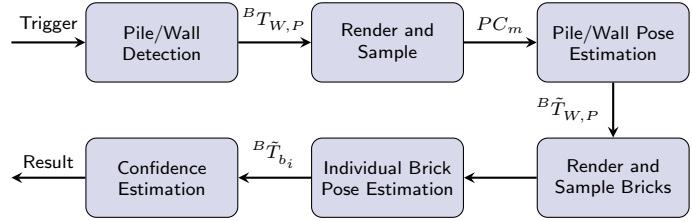
\begin{figure}[t]
	\centering\begin{maybepreview}%
	﻿%
\begin{tikzpicture}[font=\sffamily\scriptsize,on grid,>={stealth[inset=0pt,length=4pt,angle'=45]}]
\tikzset{every node/.append style={node distance=3.0cm}}
\tikzset{terminal_node/.append style={minimum size=1.5em,minimum height=3em,minimum width={width("Search Point")+0.2em},draw,align=center,rounded corners}}
\tikzset{content_node/.append style={minimum size=1.5em,minimum height=3em,minimum width={width("Search Point")+0.2em},draw,align=center,fill=blue!15!white, rounded corners}}
\tikzset{label_node/.append style={near start}}
\tikzset{group_node/.append style={align=center,rounded corners,inner sep=1em,thick}}
\tikzset{decision_node/.append style={align=center,shape aspect=1.5,minimum width=7.9em,minimum height=5.4em,diamond,draw,fill=yellow!25!white,font=\sffamily\normalsize,node distance=3.9cm}}

\definecolor{red}{rgb}     {0.5,0.0,0.0}
\definecolor{green}{rgb}   {0.0,0.5,0.0}
\definecolor{blue}{rgb}    {0.0,0.0,0.5}
\definecolor{grey}{rgb}    {0.5,0.5,0.5}

\node(marker_detection)[content_node] at (0.0, 0.0) {Pile/Wall\\Detection};
\node(render)[content_node] at (3.0,0.0){Render and\\Sample};
\node(rough_allign)[content_node] at (6.0, 0.0){Pile/Wall Pose\\Estimation};
\node(rerender)[content_node] at(6.0, -1.8){Render and\\Sample Bricks};
\node(multi_align)[content_node] at (3.0, -1.8){Individual Brick\\ Pose Estimation};
\node(conf)[content_node] at (0.0, -1.8){Confidence\\Estimation};

\draw[->, thick] (-2, 0.0) -- node[label_node,midway,above] {Trigger} (marker_detection);
\draw[->, thick] (marker_detection) -- node[label_node,midway,above] {$\T{B}{W,P}$} (render);
\draw[->, thick] (render) -- node[label_node,midway,above] {$PC_m$}(rough_allign);
\draw[->, thick] (rough_allign) -- node[label_node,midway,left] {$\TT{B}{W,P}$} (rerender);
\draw[->, thick] (rerender) -- (multi_align);
\draw[->, thick] (multi_align) -- node[label_node,midway,above] {$\TT{B}{b_i}$}(conf);
\draw[->, thick] (conf) -- node[label_node,midway,above]{Result} (-2,-1.8);

\end{tikzpicture}
\end{maybepreview}%
	\caption{LiDAR-based brick perception pipeline.}
	\label{fig:lidar_brick_perception}
\end{figure}

When the robot is close to either the pick-up location (called \textit{pile})
or the place location (called \textit{wall}), it needs to localize against
these objects and to perform pose estimation of the individual bricks in
order to pick them or place new bricks next to them. In case of an empty wall
we use our wall marker detection to locate the desired building pose (see \cref{sec:wall_marker}).

Our perception pipeline assumes knowledge of the current state of the world,
including a rough idea of the brick poses relative to the pile or wall.
The perception pipeline receives this information from the high-level control module.

Depending on the target (pile/wall, denoted as $P$/$W$), the perception pipeline receives an initial
guess of the target pose $\T{B}{P}$ or $\T{B}{W}$ w.r.t. the robot's base ($B$).
It also receives the brick pose $\T{W,P}{b_i}$ ($W,P$ denotes the considered target) and brick type $t_i \in \{\textrm{r, g, b, o}\}$
for each brick $i$.
For initial alignment purposes, the individual brick alignment can be switched off. Finally, bricks can be excluded from the optimization, for example
if they are far away and not of interest for the next action.

\Cref{fig:lidar_brick_perception} shows the overall perception process.
In both cases, it starts with a rough detection of the target location from
further away.

\subsubsection{Rough Pile Detection}

In case of the pile, we know the approximate shape beforehand. We make use
of this and search for corresponding measurements using the 3D LiDAR sensor.
While doing so, one needs to take care to disambiguate the UGV and UAV piles
and other distractors in the arena.
The first step in detecting the pile is to confine the search space to a
user-defined search area, so-called geofencing. During the competition, a rough expected location of the pile was known prior to
each run which allowed restriction of the search area. In a completely unknown scenario, the geofencing would have covered the whole arena.
We start by filtering out the points that lie outside of the search area and fit a plane to the remaining points.

\begin{figure}
	\centering\begin{maybepreview}%
	\includegraphics[clip,height=5cm,trim=0 80 0 70,frame]{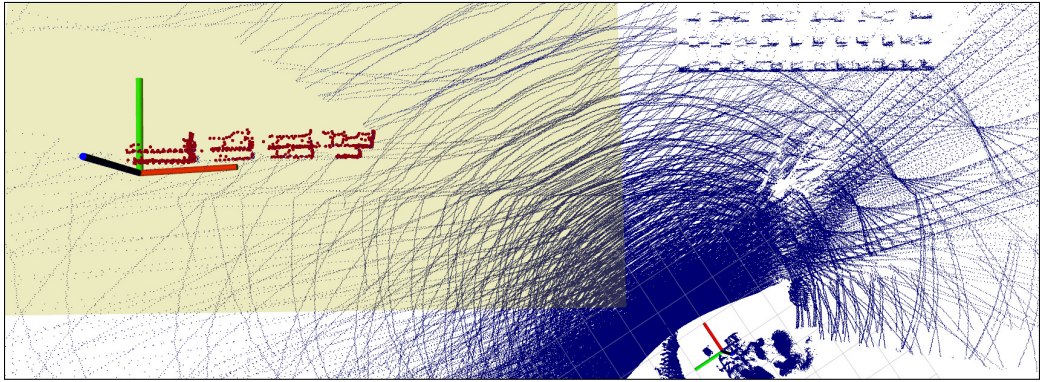}\end{maybepreview}%
	\caption{Pile detection from LiDAR points (blue).
	The robot is located at the small coordinate system (bottom).
	The search area can
	be restricted using geofencing (yellow rectangle).
	Detected points are visualized in red and the estimated pile pose is shown.
	The scan was captured inside the competition arena during a test run.
	}
	\label{fig:pile_detection}
\end{figure}

Next, we filter out the points that lie on the plane or are very close to the plane. The remaining points, shown in \cref{fig:pile_detection}, may belong to the pile.
After clustering the points and filtering clusters which do not fit the
expected pile size, we perform principal component analysis (PCA) on the remaining cluster to estimate the
largest principal component and define the pile coordinate system such that the
$X$ axis is aligned with the 2D-projected principal axis and the $Z$ axis points upwards, perpendicular to the plane.

\subsubsection{Wall Marker Detection}
\label{sec:wall_marker}
\begin{figure}
	\centering\begin{maybepreview}%
	\includegraphics[clip,height=4.5cm]{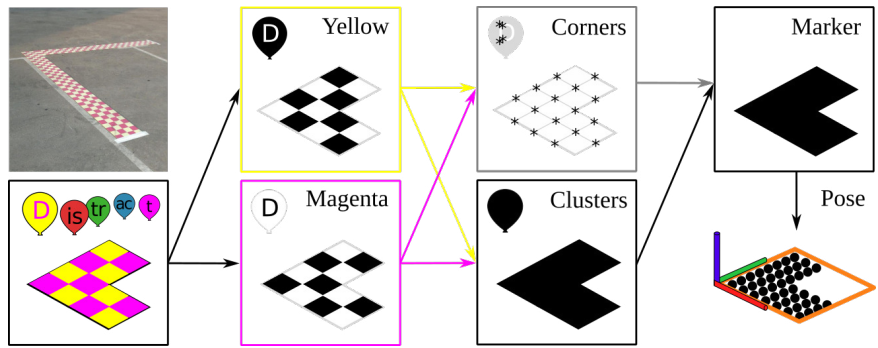}\end{maybepreview}\vspace*{-1ex}
	\caption{Wall marker detection.
	Starting from the input image (Col.~1), two color masks are generated (Col.~2). These masks are used for extracting corners (Col.~3 top) and clustering (Col.~3 bottom). Corners vote for clusters to detect the wall marker (Col.~4 top). Marker pose is estimated using oriented bounding box in orange around projected points (Col.~4 bottom).}
	\label{fig:wall_marker_detection}
\end{figure}

After picking up bricks, the next task is finding and estimating the pose of the L-shaped marker indicating where to build the wall (see \cref{fig:wall_marker_detection}). We use this perception module only if no bricks are already built. In case of a partially built wall
we localize against the 3D brick shape (see \cref{sec:rough_pile_wall}).

Our idea for detecting the marker relies on its distinctive color, pattern and shape best visible in camera images.
We use the RGB camera mounted in the endeffector for this purpose.
We start by specifying volumes within the HSV color space corresponding to the yellow and magenta tones of the marker.
Now, we exploit the characteristic color composition of the marker to filter out distractors in the image.
For that, we generate a color mask using all yellow pixels that are close to magenta pixels and another one for magenta pixels in the vicinity of yellow pixels (\cref{fig:wall_marker_detection} Col.~2).
The resulting masks preserve the pattern of the marker which we utilize to filter out further distractors.
First, we extract the corners from each mask separately and then search for those corners present in close vicinity in both masks.
Additionally, we fuse both masks and extract clusters of all masking pixels (\cref{fig:wall_marker_detection} Col.~3).
Next, we let each resulting corner vote for its corresponding cluster.
The cluster gathering most votes is assumed to be corresponding to the wall marker (\cref{fig:wall_marker_detection} Col.~4 top).

We project each cluster pixel onto the ground plane of the arena and
accumulate the resulting point clouds of the previous 10 seconds, since the camera has
limited FoV and we make use of the robot and arm movements to cover more space.
After Euclidean clustering, we compute the smallest oriented 2D rectangle around
the biggest cluster. The intersection point of the L shape can be found by
looking for the opposite corner, which should have the highest distance from
all cluster points (see \cref{fig:wall_marker_detection} Col.~4 bottom).
Finally, the detection is validated by verifying the measured side lengths.

\subsubsection{Rendering and Sampling}

The next module in the brick perception pipeline (\cref{fig:lidar_brick_perception}) converts our parametrized world model into 3D point clouds that are suitable for point-to-point registration with the measurements $PC_s$ of the Velodyne 3D LiDAR, which is moved to capture a dense 3D scan of the pile or brick scene.
We render the parametrized world model using an OpenGL-based renderer~\citep{schwarz2020stillleben}
and obtain the point cloud $PC_m$.
Both point clouds are represented in the base-link $B$.
Since we render at a high resolution of 2800$\times$2800 pixels, we downsample the
resulting point cloud to uniform density using a voxel grid filter with resolution $d$ = 0.02\,m.

\subsubsection{Rough Pile/Wall Pose Estimation}
\label{sec:rough_pile_wall}
We will now obtain a better estimate $\TT{B}{W}$ or $\TT{B}{P}$ of the pile/wall pose.
We first preprocess $PC_s$ as follows:
\begin{enumerate}
	\item Extract a cubic region around $\T{B}{W}$/$\T{B}{P}$,
	\item downsample to uniform density of using a voxel grid filter with resolution 0.02\,m,
	\item find and remove the ground plane using RANSAC, and
	\item estimate point normals (flipped such that they point towards the scanner) from local
	   neighborhoods for later usage.
\end{enumerate}
\noindent We then perform Iterative Closest Point (ICP) with a
point-to-plane cost function \citep{low2004linear} with high correspondence distance,
which usually results in a good rough alignment, followed by a point-to-point
alignment with smaller correspondence distance for close alignment.

In case the wall marker was detected, we add another cost term
\begin{equation}
 E_{dir}(\TT{B}{W}) = (1-(\RT{B}{W} \cdot (1\,0\,0)^T)^T \vec{l})^2
\end{equation}
with $\vec{l}$ being the front-line direction and $\RT{B}{W}$ the rotation component of $\TT{B}{W}$.
This cost term ensures the optimized wall coordinate system is aligned with the marker direction.

The above-defined cost function is optimized using the Ceres solver~\citep{ceres-solver} until either
the translation and rotation changes or the cost value change are below
termination thresholds ($\lambda_{T} = \num{5e-8}$, $\lambda_{C} = \num{1e-6}$).

\subsubsection{Individual Brick Pose Estimation}
\label{sec:individual_bricks}

When the robot is close enough, we can determine individual brick poses.
We constrain the following optimization to translation and yaw angle (around the
vertical $Z$ axis), since pitch and roll rotations can only happen due to
malfunctions such as dropping bricks accidentally. In these cases, the brick
will most likely not be graspable using our magnetic gripper, so we can ignore
these cases and filter them later.

For correspondence information, we re-render the scene using the
pose $\TT{B}{W,P}$ obtained from rough alignment.
Here, we include the ground plane in the rendering, since we can use it to
constrain the lowest layer of bricks.
We separate the resulting
point cloud into individual brick clouds $PC_{bj}$.

We now minimize the objective
{\small\begin{equation}
E_{\text{multi}} = \sum_{j=1}^{N}\sum_{i=1}^{M(j)}
\frac{1}{M(j)} \|(R(\theta_j) p_{j,i} + t_j - q_{j,i})^T n_{q_{j,i}}\|^2
,
\end{equation}}
where the optimized parameters $\theta_i$ and $t_i$ describe the yaw angle and translation of brick $i$,
$N$ is the number of bricks, $M(j)$ is the number of found point-to-point
correspondences for brick $j$, $p_{j,i} \in PC_{bj}$ and $q_{j,i} \in PC_{s}$
are corresponding points, and $n_q$ is the normal in point $q$.
This is a point-to-plane ICP objective with separate correspondences for each brick.
Correspondences are filtered using thresholds $\lambda_{dot}$ and $\lambda_{dist}$
for normal dot products and maximum point distances.

\begin{figure}[tb]
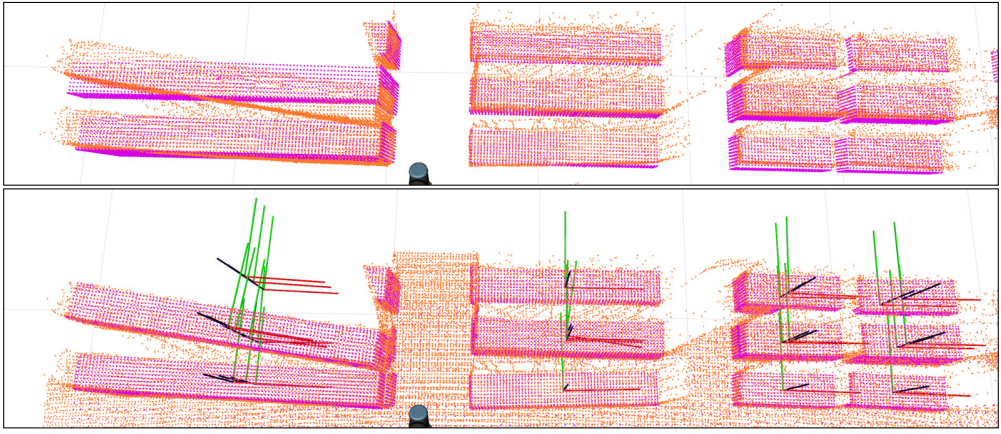

	\begin{maybepreview}%
	\begin{minipage}{\linewidth}
	\centering\includegraphics[width=0.8\linewidth,clip,trim=0 100 0 70,frame]{images/perception/pile_above_rough.jpg}\\
	\centering\includegraphics[width=0.8\linewidth,clip,trim=0 100 0 0,frame]{images/perception/pile_above_multi.jpg}
	\end{minipage}\end{maybepreview}%
	\caption{Precise alignment of individual bricks. Laser measurements are
	colored orange, model points are shown in purple.
	Top: Initial solution found by the rough ICP stage.
	Bottom: Resulting brick poses. The scan was captured during a competition run.}
	\label{fig:align}
\end{figure}

To keep the wall structure intact during optimization, we add additional cost
terms for relationships between bricks that touch each other, which punish
deviations from their relative poses in the initialization:
{\footnotesize\begin{alignat}{4}
E_{i,j}^{R} &= \lambda_r\|R(\theta_i)\R{B}{b_i}(R(\theta_j)\R{B}{b_j})^{-1}\R{B}{b_j}\R{b_i}{B} -I\|_{F}^2, \\
E_{i,j}^{T} &= \lambda_t\|t(T(\theta_i,t_i)\T{B}{b_i}(T(\theta_j,t_j)\T{B}{b_j})^{-1}\T{B}{b_j} \T{b_i}{B})\|_2^2,
\end{alignat}}
where $||\cdot||_F$ denotes the matrix norm, and $\lambda_r, \lambda_t$ are
balancing factors. Note that these pairwise cost terms have equal strength
for all involved brick pairs.

As in the rough alignment phase, the parameters are optimized using Ceres
using the same termination criteria up to a maximum of 20 iterations.
The optimization takes around 0.15\,s on the onboard computer for one iteration with 20 bricks.
As an additional output, we compute a confidence parameter for each brick as the ratio of
found correspondences to expected visible points according to the rendered model.
\Cref{fig:align} shows an exemplary result of the entire brick perception pipeline.

\subsection{Heat Source Detection}
\label{sec:thermal_detection}

The simulated fires in the ground level room of Challenge~3 are mounted inside acrylic
glass enclosures, which raises different challenges:
Firstly, only a fraction of the infrared rays penetrates the walls, so the thermal
detection works only through the hole in the front which results in a very narrow
window for detection. Secondly, the LiDAR hardly ever perceives transparent acrylic
glass, so determining the exact 3D position is another challenge.
To deal with the small detection window, we employ an arm sweeping motion and perform thresholding on the thermal images to check for heat sources.
Assuming the thermal element has reached the nominal temperature and is
observed at a nearly orthogonal angle through the hole, the bounding box of the
thermal detection can be used to estimate the distance. We used an equivalent
heat element for calibrating the measurements. Back projecting the detection's
center point with this distance results in a 3D position estimate of the heat source.

\subsection{Laser Navigation}
\label{sec:laser_navigation}

We localize our UGV w.r.t. the building using LiDAR.
In a test run, we collected data to generate a Multi-Resolution Surfel Map (MRSMap) of the arena with \citep{DavidCTSLAM}.
During localization, we register the current LiDAR point cloud against the map. Instead of estimating a correction transformation from the GPS-based \texttt{field} frame to the \texttt{map} frame, we calculate the offset transformation between the map's \texttt{field} origin and the \texttt{odometry} frame without GPS. The wheel encoder odometry is reset each time at start-up, but is accurate enough to be used as initialization during incremental scan matching.

The map is combined from two separately recorded segments, the first from the approach to the entrance and the second from
a round trip inside the simulated kitchen. Both segments are mapped as before and then aligned. This allows to center the
finer level of the multi-resolution map and thus yields a better representation where it is needed. While approaching the building,
we use the outdoor map and switch to the indoor map before passing the doorway.

\section{Evaluation}
\label{sec:evaluation}

We evaluated our UGV Bob during the MBZIRC 2020 Finals and in additional experiments in our lab.
During the MBZIRC 2020 Finals, our UGV performed in five arena runs per challenge and an additional
Grand Challenge run.
We used the three rehearsal days to get familiar with the arena conditions and
fixed Wi-Fi issues. We picked bricks in a semi-autonomous way and fine-tuned
our perception pipeline for Challenge~2. We collected LiDAR measurements of the building
and tested our thermal perception within the arena environment.

Unfortunately, in the first Challenge~2 competition run we had issues
with the gripper. We attempted over 15 times to pick up an orange
brick with very promising perception results but were unsuccessful.
We were unable to fix this problem since hardware changes were not allowed
during the competition run.
In the second competition run, we were able to autonomously pick and store a green
brick successfully, but again scored zero points since our UGV was not able to drive accurately on
the slope inside the arena to reach the wall building position. The wheel controller had an unnecessary power limit and was
tuned for flat ground. We were able to fix these problems over night. Due to limited test time,
we did not discover this problem earlier. Nevertheless, the points collected by our
UAV were enough to secure an overall second place in Challenge~2.

On the first Challenge~3 competition day, the UGV could unfortunately not compete due to software issues.
During the second challenge run the autonomy worked well to the point, where the UGV tried to position the arm for scanning the first location for heat signatures.
Because the manually configured approximate fire location was wrong (see {\cref{fig:GrandChallengeUGV} left),
the arm collided with the target which triggered the emergency stop. While trying to maneuver the robot outside during reset, the
not sufficiently tested motor controller malfunctioned and the robot drove into a cupboard.
After resetting to the start zone, the faulty controller caused the robot to drive towards the arena boundary,
even though localization and waypoints were correct. Near the end of the run, we switched to manual mode. With arm motions
and a pump macro prepared in advance, the robot managed to spray \SI{470}{\milli\litre} of water into the container just before the timeout.

After analyzing the collected data, we also found minor localization inaccuracies while entering the building (see \cref{fig:GrandChallengeUGV} left), due to proximity of inside and outside points in the map. After creating separate maps and switching on driving inside, the localization performed accurately.

\begin{figure}[h]
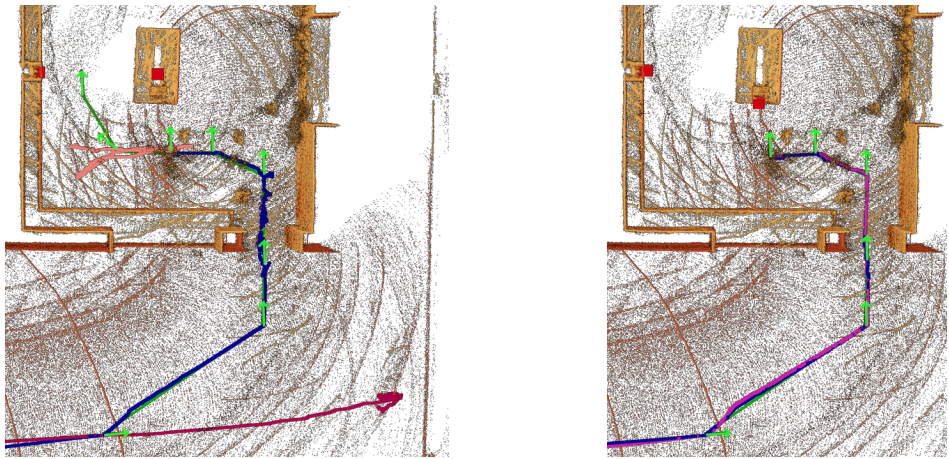

  \centering\begin{maybepreview}%
    \includegraphics[height=6cm]{images/ch3/ugv_trial_day2_2s2.png}\hspace{2cm}
    \includegraphics[height=6cm]{images/ch3/ugv_gch_fin_2s2.png}\end{maybepreview}%
  \caption{Navigation map and UGV trajectories on Trial Day~2 (left) with active Fire~2 and the Grand Challenge run (right) with active Fire~1. The light green straight lines show the planned trajectory, where each light green arrow denotes a waypoint with the direction indicating the robot's heading. The other lines are the actual driven trajectory according to localization, colored differently after each reset. Two red rectangles predefine (using the recorded map of the building) the suspected fire locations to scan for heat signatures.}
  \label{fig:GrandChallengeUGV}
\end{figure}

In the final Grand Challenge our UGV had to perform both tasks in sequence. In order to maximize our overall points, our team decided to assign
Bob the Challenge~3 task first. The UGV successfully solved the task autonomously. There was an issue with a disconnected tf-tree,
which led to the state machine stopping in front of the first fire. After a reset, the issue was solved and the UGV successfully performed as intended:
it stopped in front of the first location, scanned for the heat source,
aimed and sprayed sufficient water to achieve the full score. \cref{fig:GrandChallengeUGV} right shows driven trajectories
during the first and second attempt.
The small deviation after the first waypoint in both images was due to an edge on the uneven ground,
otherwise the overlap of planned- and driven trajectories demonstrates the high accuracy of our method.
Due to some resets caused by the UGV and the flying robots, only two minutes
were left, which was not enough time to score any points in Challenge~2.
Overall, our team NimbRo scored the second place in the MBZIRC 2020 Grand Challenge finals.

\begin{table}
\centering
\begin{threeparttable}
\caption{Build order optimization.}
\label{tab:bob_strategy}
\setlength{\tabcolsep}{6pt}
\begin{tabular}{lrrrrrr}
\toprule
Method & \multicolumn{2}{c}{$|B|$} & \multicolumn{2}{c}{$d_{B}$ [m]} & \multicolumn{2}{c}{Runtime [s]} \\
\cmidrule (lr){2-3} \cmidrule (lr){4-5} \cmidrule (lr){6-7}
       & mean & stddev & mean & stddev & mean & stddev \\
\midrule
Optimal & 5.0 & 0.91 & 3.36 & 0.97 & 7.5 & 29.0 \\
Greedy & 5.5 & 1.10 & 5.63 & 1.86 & 0.0 & 0.0 \\
\bottomrule
\end{tabular}
{\footnotesize Computing the build order ($B$) using our optimization versus a greedy approach
over 1000 randomly generated blueprints. The path length to reach all build positions is denoted as $d_B$. $|B|$ define the number of build positions given the build order $B$.}
\end{threeparttable}
\end{table}

\begin{figure}
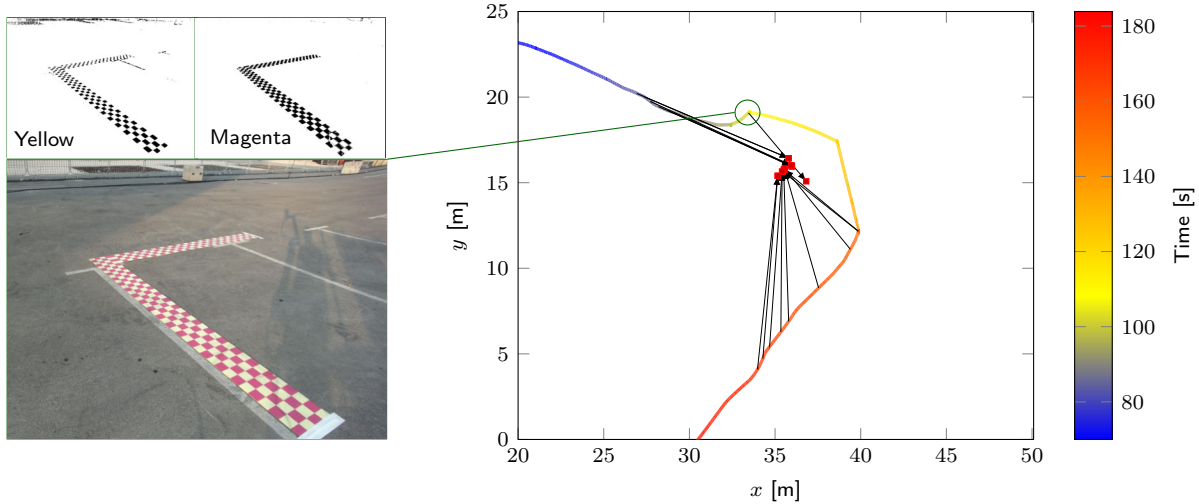
 \centering\begin{maybepreview}%
 \begin{tikzpicture}[font=\footnotesize\sffamily, line/.style={green!40!black}]
  \pgfplotstableread{data/bob_marker/marker_full.txt}\mydata
  \pgfplotstablegetrowsof{\mydata}
  \pgfmathtruncatemacro{\NumRows}{\pgfplotsretval-1}
  \begin{axis}[xlabel={$x$ [m]}, ylabel={$y$ [m]}, colorbar=true, colorbar style={ylabel={Time [s]}},
      xmin=20, ymin=0, ymax=25,axis equal=true,
      point meta min=70]
    \addplot+[no marks,mesh,very thick] table [x=x, y=y, point meta=\thisrow{time}] {data/bob_marker/robot_full.txt};
    \pgfplotsinvokeforeach{0,...,\NumRows}{ %
        \pgfplotstablegetelem{#1}{x}\of{\mydata}\edef\X{\pgfplotsretval}
        \pgfplotstablegetelem{#1}{y}\of{\mydata}\edef\Y{\pgfplotsretval}
        \pgfplotstablegetelem{#1}{rx}\of{\mydata}\edef\RX{\pgfplotsretval}
        \pgfplotstablegetelem{#1}{ry}\of{\mydata}\edef\RY{\pgfplotsretval}

        \edef\tmp{\noexpand\draw[-latex,very thin] (axis cs:\RX,\RY) -- (axis cs:\X,\Y);}\tmp
    }
    \addplot+[only marks, mark size=1pt] table [x=x, y=y] {\mydata};
  \end{axis}
  \node[anchor=south west,inner sep=0, draw=green!40!black, thick] (rgb) at (-6.8,0) {\includegraphics[clip,trim=250 0 0 0, height=3.7cm]{images/experiments/marker_detection/rgb1.png}};
  \node[anchor=south west,inner sep=0, draw=green!40!black, thick] (image) at (-6.8,3.74) {\includegraphics[clip,trim=250 0 0 0, height=1.85cm]{images/experiments/marker_detection/yellow1.png}};
  \node[anchor=south west,inner sep=0, draw=green!40!black, thick] (image) at (-4.3,3.74) {\includegraphics[clip,trim=250 0 0 0, height=1.85cm]{images/experiments/marker_detection/magenta1.png}};
  \node[circle, draw=green!40!black](circle) at (3.05, 4.35){};
  \draw[line](rgb.north east)--(circle.west);
  \node[rectangle, inner sep=0.5, anchor=center] at(-6.3,4.0){Yellow};
  \node[rectangle, inner sep=0.5, anchor=center] at(-3.55,4.0){Magenta};
 \end{tikzpicture}\end{maybepreview}%
 \caption{Marker detection during practice run. The robot was teleoperated through the arena, driving past the marker. Bob's trajectory is shown colored by time. Each detection is shown as a red square, with an arrow from
 Bob's position at detection time. The images on the left show an example scene captured at the green circle including the yellow and magenta color mask output.}
 \label{fig:eval:marker}
\end{figure}

After the competition, we evaluated the wall marker detection (see \cref{sec:wall_marker}) on a dataset recorded inside the arena during a
rehearsal run. The robot was teleoperated through the arena, driving past the marker.
There is a gap of 45 seconds in the marker predictions after the spot marked with a green circle,
since Bob was not facing the marker while driving to the arena boundary.
The outlier in the detected marker position can be explained by the slope the robot was driving on. First, our localization using only wheel odometry
had problems when driving over the edges at the beginning and end of the slope. Second, we assumed the arena would be placed on a flat area.
The height increase relative to the robot starting pose led to an imprecise projection of the estimated marker onto the expected ground plane.
Nevertheless, the results show that the marker detection worked reliably in the challenging lighting conditions with a distance
of up to 11\,m between the robot and the marker. The standard deviation of the marker position relative to robot starting location was 0.40\,m (in the $x$ direction of our arena frame) and 0.35\,m ($y$ direction). Ignoring the outlier generated by the slope, the standard deviation drops to 0.3\,m in both directions.
Since we fixed the motor controller to handle the slope before the Grand Challenge run,
this would have been robust enough to locate the wall marker during the competition.

In addition to the competition, we evaluated two sub-systems and a simplified Challenge~2 task in
our lab environment.

We compared our algorithm for optimizing the build order (see \cref{sec:ch2fsm}) with a greedy strategy.
Using the greedy strategy, we take the best local solution, i.e. given a set of already built
bricks, we chose the next build position such that the number of bricks the UGV able to build
is maximal. \cref{tab:bob_strategy} shows the results of both approaches on
1,000 randomly generated blueprints. The optimization reduces the different build positions
needed from 5.5 to 5.0 on average and gives an even larger improvement regarding the
distance needed to be driven by 2.3\,m on average. Performing the optimization
takes on average 7.5\,s, which is feasible in our use case since it is performed just once before the
competition run. Nevertheless, it is---as expected---much slower than the greedy approach
due to the exponential complexity.

\begin{figure}[b]
 \centering\begin{maybepreview}%
 ﻿\resizebox{.95\linewidth}{!}{%
\begin{tikzpicture}
[font=\sffamily\Large]

\coordinate (r) at (18.5,0);
\coordinate (g) at (12.5,3);
\coordinate (b) at (0,3);
\coordinate (o) at (0,0);
\coordinate (l) at (20,3.7);

\draw[thick, black] (r) rectangle ($(r) + (3,2)$);
\draw[thin, red, rotate around={0.95:($(r) + (0.17,0.24)$)}] ($(r) + (0.17,0.24)$) rectangle ($(r) + (3.17,2.24)$);
\draw[thin, red, rotate around={1.72:($(r) + (0.11,0.17)$)}] ($(r) + (0.11,0.17)$) rectangle ($(r) + (3.11,2.17)$);
\draw[thin, red, rotate around={1.91:($(r) + (0.17,0.23)$)}] ($(r) + (0.17,0.23)$) rectangle ($(r) + (3.17,2.23)$);
\draw[thin, red, rotate around={0.38:($(r) + (0.16,0.11)$)}] ($(r) + (0.16,0.11)$) rectangle ($(r) + (3.16,2.11)$);
\draw[thin, red, rotate around={0:($(r) + (0.19,0.2)$)}] ($(r) + (0.19,0.2)$) rectangle ($(r) + (3.19,2.2)$);
\draw[thin, red, rotate around={-1.34:($(r) + (0.1,0.1)$)}] ($(r) + (0.1,0.1)$) rectangle ($(r) + (3.1,2.1)$);
\draw[thin, red, rotate around={-0.38:($(r) + (0.13,0.11)$)}] ($(r) + (0.13,0.11)$) rectangle ($(r) + (3.13,2.11)$);
\draw[thin, red, rotate around={2.29:($(r) + (0.19,0.2)$)}] ($(r) + (0.19,0.2)$) rectangle ($(r) + (3.19,2.2)$);
\draw[thin, red, rotate around={1.53:($(r) + (0.18,0.18)$)}] ($(r) + (0.18,0.18)$) rectangle ($(r) + (3.18,2.18)$);
\draw[thin, red, rotate around={2.1:($(r) + (0.17,0.19)$)}] ($(r) + (0.17,0.19)$) rectangle ($(r) + (3.17,2.19)$);

\draw[thick, black] (g) rectangle ($(g) + ((6,2)$);
\draw[thin, green, rotate around={-0.95:($(g) + (0.21,0.04)$)}] ($(g) + (0.21,0.04)$) rectangle ($(g) + (6.21,2.04)$);
\draw[thin, green, rotate around={2.01:($(g) + (0.14,0.16)$)}] ($(g) + (0.14,0.16)$) rectangle ($(g) + (6.14,2.16)$);
\draw[thin, green, rotate around={-1.05:($(g) + (0.2,-0.04)$)}] ($(g) + (0.2,-0.04)$) rectangle ($(g) + (6.2,1.96)$);
\draw[thin, green, rotate around={-0.95:($(g) + (0.2,0.04)$)}] ($(g) + (0.2,0.04)$) rectangle ($(g) + (6.2,2.04)$);
\draw[thin, green, rotate around={1.72:($(g) + (0.16,0.16)$)}] ($(g) + (0.16,0.16)$) rectangle ($(g) + (6.16,2.16)$);
\draw[thin, green, rotate around={0.76:($(g) + (0.13,0.05)$)}] ($(g) + (0.13,0.05)$) rectangle ($(g) + (6.13,2.05)$);
\draw[thin, green, rotate around={-2.67:($(g) + (0.23,-0.01)$)}] ($(g) + (0.23,-0.01)$) rectangle ($(g) + (6.23,1.99)$);
\draw[thin, green, rotate around={-1.05:($(g) + (0.2,-0.04)$)}] ($(g) + (0.2,-0.04)$) rectangle ($(g) + (6.2,1.96)$);
\draw[thin, green, rotate around={-0.38:($(g) + (0.19,0.1)$)}] ($(g) + (0.19,0.1)$) rectangle ($(g) + (6.19,2.1)$);
\draw[thin, green, rotate around={1.05:($(g) + (0.16,0.13)$)}] ($(g) + (0.16,0.13)$) rectangle ($(g) + (6.16,2.13)$);

\draw[thick, black] (b) rectangle ($(b) + (12,2)$);
\draw[thin, blue, rotate around={0.57:($(b) + (0.21,0.12)$)}] ($(b) + (0.21,0.12)$) rectangle ($(b) + (12.21,2.12)$);
\draw[thin, blue, rotate around={0.67:($(b) + (0.1,0.1)$)}] ($(b) + (0.1,0.1)$) rectangle ($(b) + (12.1,2.1)$);
\draw[thin, blue, rotate around={0.91:($(b) + (0.17,0.165)$)}] ($(b) + (0.17,0.165)$) rectangle ($(b) + (12.17,2.165)$);
\draw[thin, blue, rotate around={0.93:($(b) + (0.17,0.17)$)}] ($(b) + (0.17,0.17)$) rectangle ($(b) + (12.17,2.17)$);
\draw[thin, blue, rotate around={1.1:($(b) + (0.26,0.18)$)}] ($(b) + (0.26,0.18)$) rectangle ($(b) + (12.26,2.18)$);
\draw[thin, blue, rotate around={0.91:($(b) + (0.13,0.17)$)}] ($(b) + (0.13,0.17)$) rectangle ($(b) + (12.13,2.17)$);
\draw[thin, blue, rotate around={0.91:($(b) + (0.22,0.15)$)}] ($(b) + (0.22,0.15)$) rectangle ($(b) + (12.22,2.15)$);
\draw[thin, blue, rotate around={1.38:($(b) + (0.16,0.16)$)}] ($(b) + (0.16,0.16)$) rectangle ($(b) + (12.16,2.16)$);
\draw[thin, blue, rotate around={0.05:($(b) + (0.165,0.06)$)}] ($(b) + (0.165,0.06)$) rectangle ($(b) + (12.165,2.06)$);
\draw[thin, blue, rotate around={0.19:($(b) + (0.22,0.05)$)}] ($(b) + (0.22,0.05)$) rectangle ($(b) + (12.22,2.05)$);

\draw[thick, black] (o) rectangle ($(o) + (18,2)$);
\draw[thin, orange, rotate around={-0.38:($(o) + (0.11,-0.07)$)}] ($(o) + (0.11,-0.07)$) rectangle ($(o) + (18.11,1.93)$);
\draw[thin, orange, rotate around={-0.32:($(o) + (0.16,-0.07)$)}] ($(o) + (0.16,-0.07)$) rectangle ($(o) + (18.16,1.93)$);
\draw[thin, orange, rotate around={-0.35:($(o) + (0.15,-0.04)$)}] ($(o) + (0.15,-0.04)$) rectangle ($(o) + (18.15,1.96)$);
\draw[thin, orange, rotate around={0.45:($(o) + (0.12,0.14)$)}] ($(o) + (0.12,0.14)$) rectangle ($(o) + (18.12,2.14)$);
\draw[thin, orange, rotate around={0.73:($(o) + (0.05,0.17)$)}] ($(o) + (0.05,0.17)$) rectangle ($(o) + (18.05,2.17)$);
\draw[thin, orange, rotate around={0.22:($(o) + (0.16,0.02)$)}] ($(o) + (0.16,0.02)$) rectangle ($(o) + (18.16,2.02)$);
\draw[thin, orange, rotate around={-0.92:($(o) + (0.22,-0.09)$)}] ($(o) + (0.22,-0.09)$) rectangle ($(o) + (18.22,1.91)$);
\draw[thin, orange, rotate around={0.54:($(o) + (0.05,0.17)$)}] ($(o) + (0.05,0.17)$) rectangle ($(o) + (18.05,2.17)$);
\draw[thin, orange, rotate around={0.13:($(o) + (0.14,0.07)$)}] ($(o) + (0.14,0.07)$) rectangle ($(o) + (18.14,2.07)$);
\draw[thin, orange, rotate around={0.45:($(o) + (0.16,0.14)$)}] ($(o) + (0.16,0.14)$) rectangle ($(o) + (18.16,2.14)$);

\node (x) at ($(l) + (1.1,0)$){x} ;
\node (y) at ($(l) + (0,1.1)$){y} ;
\draw[->, line width=0.3mm](l)  -- ($(l) + (0,0.8)$);
\draw[->, line width=0.3mm](l)  -- ($(l) + (0.8,0)$);

\end{tikzpicture}
}
\end{maybepreview}\vspace*{-2ex}
 \caption{UGV pick and place robustness. Ground truth place pose (black) and ten test results for each brick type.}
 \label{fig:bob_eval_grasping}
\end{figure}
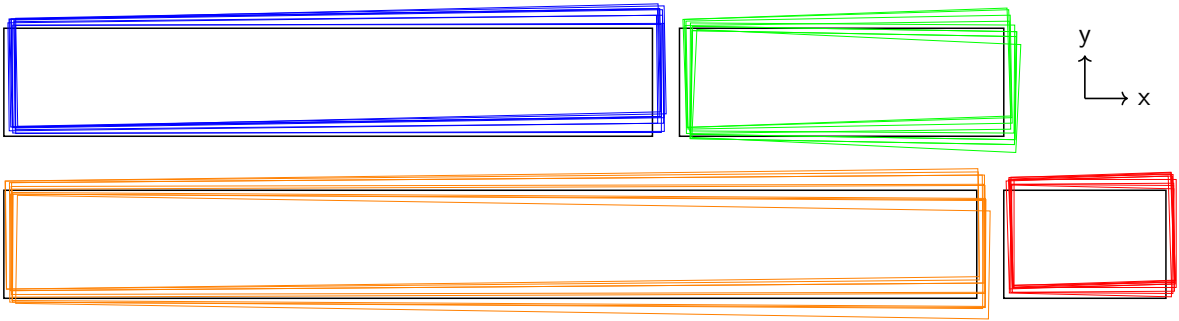

In a second lab experiment, we evaluated the precision and repeatability of
picking up bricks from a pile (see \cref{fig:bob_eval_grasping}). We placed four bricks in front of the robot
in a similar way the piles in the competition were arranged. The target place pose was on the left side of the robot on the floor.
Each test consists of scanning the piles, picking the brick with the highest confidence, and placing it
at the target pose. We than calculated the mean translation and rotation error
compared to the ground truth pose, which was defined by placing a perfectly center-grasped brick at the target location.
Only a rough estimation of the pile
location was provided to the system. We repeated the test ten times for each brick type while changing the brick
horizontal positions by up to 5\,cm and the rotation by up to 10$^\circ$ around the vertical axis.
\Cref{tab:bob_grasping} shows the mean results per brick for ten tests each.
The resulting mean error could be further decreased by investing more time calibrating
the whole system; nevertheless, it is sufficient to place the bricks reliably
into the storage system. The very low standard deviation in both rotation and translation
shows that our perception and grasping pipeline has a high repeatability and is very
robust.

\begin{table}
\centering
\begin{threeparttable}
 \caption{End-to-end brick manipulation precision.}
\label{tab:bob_grasping}
\setlength{\tabcolsep}{6pt}
\begin{tabular}{lrrrrrr}
\toprule
Brick & \multicolumn{2}{c}{Translation x [cm]} & \multicolumn{2}{c}{Translation y [cm]} & \multicolumn{2}{c}{Yaw [$^\circ$]} \\
\cmidrule (lr) {2-3} \cmidrule (lr){4-5} \cmidrule (lr){6-7}
      & mean & stddev & mean & stddev & mean & stddev \\
\midrule
Red    & 1.52 & 0.36 & 1.49 & 0.37 & 1.26 & 0.80 \\
Green  & 1.94 & 0.33 & 0.67 & 0.43 & 1.26 & 0.68 \\
Blue   & 1.82 & 0.49 & 0.53 & 0.21 & 0.76 & 0.40 \\
Orange & 1.34 & 0.65 & 0.36 & 0.39 & 0.45 & 0.24 \\
\bottomrule
\end{tabular}
{\footnotesize Placement error from perceiving, picking, and placing. Ten tests per color.}
\end{threeparttable}
\end{table}

\begin{table}
 \caption{Autonomous wall building test results.}
\label{tab:bob_expPipeline}
\centering
\setlength{\tabcolsep}{6pt}
\adjustbox{valign=T}{
\begin{tabular}{cccc}
\toprule
Run & Success & Time [min] & Reason of failure\\
\midrule
1 & yes & 10:40 & \\
2 & yes & 10:45 & \\
3 & no  & - & storage collision \\
4 & no  & - & arm protective stop\\
5 & yes & 11:06 & \\
6 & yes & 11:01 & \\
\bottomrule
\end{tabular}
}\hspace{0.5cm}
\adjustbox{valign=T}{
\begin{tabular}{lc}
\toprule
Component & avg. Time [min]\\
\midrule
High-level planner & 0:03\\
Pile perception & 1:05\\
Pile locomotion & 0:50\\
Pile manipulation & 5:00\\
Wall perception & 0:30\\
Wall locomotion & 0:17\\
Wall manipulation & 3:06\\
\midrule
Full system & 10:53\\
\bottomrule
\end{tabular}
}\\
\vspace{0.2cm}
{\footnotesize Six autonomous test runs evaluating the whole system. 11 bricks had to be picked, transported and placed. }
\end{table}

In addition to the sub-system evaluations, we tested our whole pipeline with a simplified Challenge~2 task (see \cref{fig:expOverview}).
The lab experiments were performed on the smaller Mario platform (UGV winner of MBZIRC 2017) with a down-scaled version of the storage system. Mario has been equipped with exactly the same drive system and arm as Bob.
The main difference between both robots is their footprint (1.9$\times$1.4\,m versus  1.08$\times$0.78\,m).
Thus, Mario is only capable to carry a single storage compartment. We modified the high-level planner to
support multiple bricks of different types in the smaller storage system. The remaining components are the same compared to Bob.
To test the full pipeline, we placed three blue, four green, and four red bricks in piles
similar to the competition conditions. The visual marker for the desired wall pose
was placed approx. 4\,m apart from the pile location. Six evaluation runs were performed,
each consisting of autonomous loading, transporting, and placing 11 bricks. Only rough pile and wall locations were provided to the robot.
The results can be seen in \cref{tab:bob_expPipeline}. Four out of six runs were fully successful. The UGV needed around 11\,minutes to solve the
task. As reported in \cref{tab:bob_expPipeline}, the manipulation actions consume most of the time. In contrast, locomotion and planning time is negligible.
During Run~3, the robot decided to load the bricks in a different order than expected. Due to the different arrangement of the storage system
on the smaller Mario robot, the arm collided with the storage system. After fixing this issue, the UR10e arm stopped during Run~4 while moving a brick due to communication problems with the controlling PC. The implemented arm recovery strategy worked as expected but the run failed since the robot lost the brick during the stop and needed all bricks to
fully complete the task. The UGV performed as expected in the remaining four runs resulting in similar execution times, which shows the robustness of the system. Two example walls built fully autonomously can be seen in \cref{fig:expPipeline}.

\begin{figure}[h]
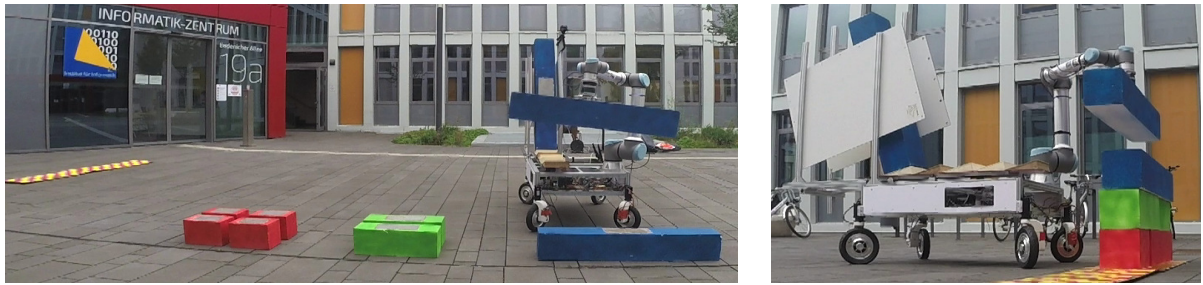

  \centering\begin{maybepreview}%
    \includegraphics[clip,trim=0 0 30 0, height=3.7cm]{images/experiments/pick_crop.png}\hspace{.02\linewidth}
    \includegraphics[clip,trim=0 0 0 0, height=3.7cm]{images/experiments/place_crop.png}\end{maybepreview}%
    \caption{Autonomous wall building. Bob is loading bricks (left) and building the predefined wall (right).}
  \label{fig:expOverview}
\end{figure}

\begin{figure}[h]
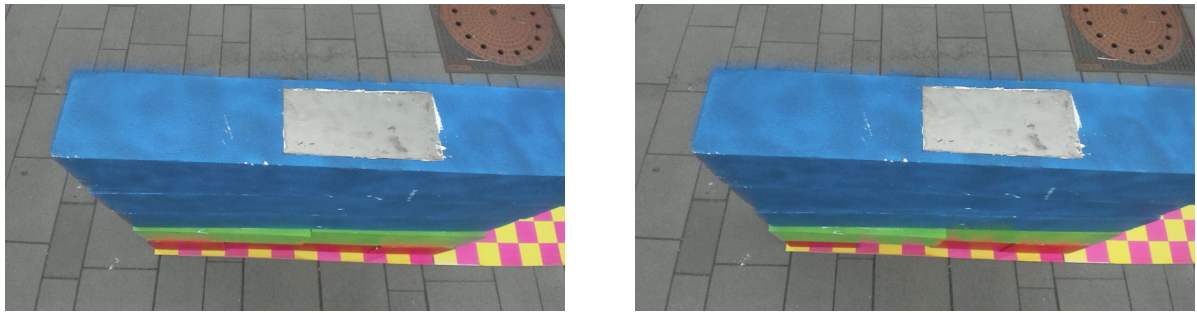

  \centering\begin{maybepreview}%
    \includegraphics[clip,trim=100 0 0 70, width=.45\linewidth]{images/experiments/image.png}\hspace{.05\linewidth}
    \includegraphics[clip,trim=100 0 0 70, width=.45\linewidth]{images/experiments/image2.png}\end{maybepreview}%
    \caption{Experiment: Two example results building a predefined wall structure fully autonomously. Tiny differences in the brick alignment can be seen
  at the green bricks.}
  \label{fig:expPipeline}
\end{figure}

\section{Lessons Learned and Conclusion}
\label{sec:Lessons_Learned}

We take the opportunity to identify key strengths and weaknesses of our system
and development approach. We also want to identify aspects of the competition
that could be improved to increase scientific benefit in the future.

Regarding our system, we saw very little problems with our hardware design itself.
For Challenge~2, reducing the number of required trips between pile and wall, Bob could have scored a large number of points in little time.
After solving initial problems with our magnets, it could manipulate even the large orange bricks.
While the large distance between pile and wall is challenging, our system is not reliant on precise global navigation. Instead we react to local pose estimates for pile, bricks, and wall, so that inaccuracies cannot accumulate.
In Challenge~3, the distance between the robot start and target location was much smaller. In addition, we used global
LiDAR localization which ensured precise locomotion. However,
the sloped terrain posed unexpected problems for our drive system and controller.

The biggest issue shortly before and during the competition was unavailable
testing time in comparison to the complexity of the developed system.
Due to resource conflicts when building all robots needed for the MBZIRC 2020
competition, hardware and software component development and testing in our lab was
executed on the modified Mario robot which was slightly different
from the Bob robot used at the competition.
Robust solutions require full-stack testing under competition
constraints. Since we postponed many design decisions until the rules were settled,
our complex design could not be tested fully before the event.
In hindsight, simpler designs with fewer components, which would have
required less thorough testing, could have been more successful in the short available time frame.

Notable to all teams, the MBZIRC 2020 suffered from general low team performance.
We think there are underlying systematic causes for this low team performance.
From the perspective of participants, we think the frequent and late changes of the rules have certainly contributed to this situation.
A pre-competition event such as the Testbed in the DARPA Robotics Challenge
can help to identify key issues with rules and material early in the competition
time-line.
Also, MBZIRC 2020 required participants to address seven completely different tasks if they wanted to gain maximum points,
ideally with custom-tailored robots for each task. Maybe focusing on general usability by constraining
the number of allowed platforms per team could lower the required effort and allow smaller teams to compete as well.

On the positive side, we noticed that having back-up strategies to address the task on a lower level of autonomy
is highly useful. Such assistance functions increase flexibility and can make the difference between not solving the task at all and
solving it partially---as demonstrated during our Challenge~3 run which scored manual points.
Also reducing high-level control to a bare minimum, instead of a universal strategy, makes the systems easier to test and reduces potential errors.

In the future, we will increase our focus on testing the developed components in competition-related scenarios to increase the robustness of
the system and to not depend on extensive testing on site during the rehearsal days.
Regarding the locomotion system, we would like to increase the locomotion
precision even further by incorporating IMU and---if available---GPS data.

In this article, we presented our UGV Bob developed for two challenges at the MBZIRC 2020 competition. The robot successfully delivered autonomously the maximum amount of water
into the fire in the Grand Challenge. Even though this robot was not able to score any points in Challenge~2 during the competition, after analyzing and fixing the issues,
our experiments show that the system functions reliably. The perception and manipulation modules worked reliably and accurately. Our developed hard- and software systems for the MBZIRC 2017 competition together with the gained experience has proven to be useful for developing related systems. Similarly, the knowledge gained and components developed in MBZIRC 2020 will also
prove useful in future applications.

\subsubsection*{Acknowledgments}
We would like to thank all members of our team NimbRo for their support before and during the competition.
This work has been supported by a grant of the Mohamed Bin Zayed International Robotics Challenge (MBZIRC).

\bibliographystyle{apalike}
\bibliography{references}

\end{document}